\documentclass{svjour3}                     
\smartqed  
\usepackage{graphicx}
\usepackage{amssymb}
\usepackage{amsfonts}

\usepackage{array}
\usepackage{slashbox}
\usepackage[cmex10]{amsmath}
\usepackage{array}
\usepackage{stfloats}
\usepackage{url}

\usepackage{comment}
\usepackage{natbib}

\begin{document}

\title{On Tsallis Entropy Bias and Generalized
Maximum Entropy Models%
}


\author{Yuexian Hou         \and
        Tingxu Yan  \and
        Peng Zhang  \and
        Dawei Song  \and
        Wenjie Li 
}


\institute{Yuexian Hou and Tingxu Yan \at
              School of Computer Science and Technology, Tianjin University, Tianjin, 300072, China \\
              \email{yxhou@tju.edu.cn; Sunriser2008@gmail.com}
           \and
           Peng Zhang and Dawei Song \at
              School of Computing, The Robert Gordon University, Aberdeen, AB25 1HG, United
              Kingdom \\
              \email{\{p.zhang1;\ d.song\}@rgu.ac.uk}
           \and
           Wenjie Li \at
              Department of Computing, The Hong Kong Polytechnic University, Hong Kong, 999077 \\
              \email{cswjli@comp.polyu.edu.hk}
}
\date{Received: date / Accepted: date}

\maketitle

\begin{abstract}
In density estimation task, maximum entropy model (Maxent) can
effectively use reliable prior information via certain constraints,
i.e., linear constraints without empirical parameters. However,
reliable prior information is often insufficient, and the selection
of uncertain constraints becomes necessary but poses considerable
implementation complexity. Improper setting of uncertain constraints
can result in overfitting or underfitting. To solve this problem, a
generalization of Maxent, under Tsallis entropy framework, is
proposed. The proposed method introduces a convex quadratic
constraint for the correction of (expected) Tsallis entropy bias
(TEB). Specifically, we demonstrate that the expected Tsallis
entropy of sampling distributions is smaller than the Tsallis
entropy of the underlying real distribution. This expected entropy
reduction is exactly the (expected) TEB, which can be expressed by a
closed-form formula and act as a consistent and unbiased correction.
TEB indicates that the entropy of a specific sampling distribution
should be increased accordingly. This entails a quantitative
re-interpretation of the Maxent principle. By compensating TEB and
meanwhile forcing the resulting distribution to be close to the
sampling distribution, our generalized TEBC Maxent can be expected
to alleviate the overfitting and underfitting. We also present a
connection between TEB and Lidstone estimator. As a result,
TEB-Lidstone estimator is developed by analytically identifying the
rate of probability correction in Lidstone. Extensive empirical
evaluation shows promising performance of both TEBC Maxent and
TEB-Lidstone in comparison with various state-of-the-art density
estimation methods.

\keywords{Density estimation \and Maximum entropy \and Tsallis
entropy \and Tsallis entropy
  bias \and Lidstone estimator}
\end{abstract}

\section{Introduction}

\label{intro} The maximum entropy (Maxent) approach to density
estimation was originally proposed by E. T. Jaynes~\citep{Jaynes57},
and since then has been widely used in many areas of computer
science and statistical learning, especially natural language
processing~\citep{Berger96,Pietra97}. The Maxent principle can be
traced back to Jaynes' classical description~\citep{Jaynes57}:

``$\ldots$ the fact that a probability distribution maximizes
entropy subject to certain constraints representing our incomplete
information, is the fundamental property which justifies use of that
distribution for inference; it agrees with everything that is known,
but carefully avoids assuming anything that is not known$\ldots$''

In implementing this principle, given a sampling distribution drawn
from the underlying real distribution, Maxent computes a resulting
distribution whose entropy is maximized, subject to a set of
selected constraints. The standard Maxent can be formulated in
Formula~\ref{eq:sm}:

\begin{equation}\label{eq:sm}
\begin{split}
& \max_{\bar{P}^{(m)}} \ S[\bar{P}^{(m)}] \\
& st. \ \ \ | \sum_{j \in S_u}{\bar{p}_j} - a_u | \leq  \delta_u \ \ \ \ \ \ \forall u \in U\\
& \ \quad \left\{ \begin{aligned}
\sum_{i \in S_c}\bar{p}_i = a_c^{*} \ \ or, \ \ \ \ \ \ \ \ \ \ \  \quad \quad \quad \quad \\
\sum_{i \in S_c}\bar{p}_i \geq a_c^{*} \ \ or,  \ \  \ \ \  \quad \forall c \in C \quad \quad \\
\sum_{i \in S_c}\bar{p}_i \leq a_c^{*} \ \ \ \ \ \ \ \ \ \ \  \quad
\quad  \quad \quad \quad \quad
\end{aligned} \right. \\
\end{split}
\end{equation}
where ${\bar P}^{\left( m \right)} \equiv \left\langle {{\bar p}_1 ,
\ldots ,{\bar p}_m } \right\rangle $ is the resulting $m$-nomial
probability distribution, $S[\cdot]$ denotes the Shannon entropy of
some probability distribution, $C$ and $U$ are two index sets,
$a_c^{*}$, $c \in C$ is the constant determined by reliable
information, $a_u$ and $\delta_u$, $u \in U$ are the parameters that
need to be empirically adjusted, and $S_c, c \in C$ and $S_u, u \in
U$ are subsets of $\{1,2,\ldots, m\}$. Standard Maxent has two sets
of constraints. The first set (indexed by $C$) includes all certain
constraints, which are derived from reliable prior information and
do not involve empirical parameters. For example, two of the most
common certain constraints are $\sum_i{\bar{p}_i}=1$ and $\bar{p}_i
\geq 0$. The second set (indexed by $U$) includes all uncertain
constraints, which are from less reliable knowledge or sample
information, and hence necessarily involve empirical parameters
(e.g., $a_u$ and $\delta_u$) to gain a satisfying performance. Note
that there could be other specific forms of constraints not listed
in Formula~\ref{eq:sm}, e.g., the common form of real-valued feature
functions. However, these constraint forms are essentially
equivalent to and can be categorized into certain or uncertain
constraints. Moreover, all certain and uncertain constraints
considered in this paper are linear.

\subsection{The Problem}
Although the essential idea of Maxent is concise and elegant, the
implementation of Maxent poses considerable practical complexity.
Specifically, in a typical density estimation task, reliable prior
information is often insufficient. In this case, if Maxent only
involves certain constraints derived from reliable prior
information, the resulting distribution will be away from the
sampling distribution. Consequently, underfitting will result.
Hence, Maxent usually involves a set of uncertain constraints, which
force the resulting distribution to be close to the sampling
distribution. The tolerable violation-level of the resulting
distribution against the sampling distribution is controlled by a
set of threshold parameters. These constraints and parameters are
essentially empirical and ad-hoc. This is a dilemma: On one hand, if
a large number of uncertain constraints and a set of tight threshold
parameters are involved, the solution of Maxent will be close to the
sampling distribution and might severely overfit the
sample~\citep{Miroslav07}; On the other hand, if a small number of
uncertain constraints or a set of loose threshold parameters are
used, Maxent might underfit the sample and miss out some useful
sample information.

\subsection{Existing Work} In the framework of Maxent, main
approaches to tackling overfitting or underfitting are parameter
regularization and constraint relaxation. The former introduces some
specific statistics (e.g., $l_1$, $l_2^2$, $l_1+l_2^2$ etc.) as the
regularized terms of the objective function and removes explicit
constraints~\citep{Miroslav07,Chen00,Lebanon01,Lau94}. The latter
aims to relax the constraints according to some theoretical
considerations ~\citep{Khudanpur95,Kazama03,Jedynak05}, e.g.,
Maximum Likelihood set in ~\citep{Jedynak05}. The performance
guarantee of some Maxent variants is rigorously established with
respect to (w.r.t.) finite sample criteria, e.g., Probably
Approximately Correct (PAC). However, according to our best
knowledge, most of guarantees are, to some extent, self-referencing.
For example, using log loss as the criterion, theoretical relations
between the solution of Generalized Maxent (GME) and the best Gibbs
distribution are given~\citep{Miroslav07}. However, the definition
of the best Gibbs distribution intrinsically depends on the
selection of feature functions. It turns out that, if the selection
of feature functions is improper, the solution of GME might not be
able to avoid overfitting or underfitting substantially even if it
is close to the best Gibbs distribution.

\subsection{Our Approach}
In this paper, we propose a novel generalization of Maxent, under
the framework of Tsallis entropy\footnote{ Please refer to
Section~\ref{def} for more details about Tsallis entropy. For the
sake of analytical simplicity, we only consider the Tsallis entropy
with $q = 2$~\citep{Tsallis88} in this
paper.}~\citep{Tsallis88,Abe00}.

An important motivating observation is that, the expected Tsallis
entropy of sampling distributions is always smaller than the Tsallis
entropy of the underlying real distribution. To demonstrate this
formally, we present a theoretical analysis on the expected Tsallis
entropy bias (TEB)\footnote{In this paper, the notation of ``Tsallis
entropy bias'' has the same meaning as the ``expected Tsallis
entropy bias''. Accordingly, TEB has the ``expected'' sense in
itself.} between sampling distributions and the underlying real
distribution. The TEB is independent of the
selection\footnote{Actually, the TEB only depend on i.i.d. sampling
presumption.} of constraints and can be expressed by a simple
closed-form formula of the sample size $n$ and the Tsallis entropy
of the underlying real distribution. This observation naturally
entails a quantitative re-interpretation and a theoretical guarantee
of the Maxent principle: Since the entropy of sampling distributions
is smaller, in the expected sense, than the entropy of the
underlying real distribution, Maxent should increase the entropy of
the sampling distribution to compensate the TEB and hence
approximate the underlying real distribution. The TEB is first
developed in the frequentist framework and we notate it as
Frequentist-TEB. In addition, by assuming a uniform Bayesian prior
over all possible $m$-nomial distributions, a Bayesian-TEB is
developed.

We argue that, in consistency with the basic principle of Maxent, a
rigorously established compensation of Frequentist-TEB or
Bayesian-TEB can help alleviate the overfitting problem. On the
other hand, it is natural to overcome underfitting through simply
forcing the resulting distribution to be close to the sampling
distribution. By integrating these two strategies into our
generalized Maxent, called Tsallis entropy bias compensation (TEBC)
Maxent, it is expected that TEBC Maxent can alleviate overfitting
and underfitting. Note that it is somewhat problematic to develop
the similar method in the framework of Shannon entropy since a
consistent and unbiased correction of Shannon entropy has not been
exactly found yet, in general (see Section~\ref{relworks} for more
detials of the estimate of Shannon entropy)

In implementation, the TEBC Maxent is convex and hence can be
efficiently solved. More importantly, TEBC Maxent can bypass the
selection of uncertain constraints as well as parameter
identification by introducing a parameter-free TEB constraint,
aiming at quantitative entropy compensation.

In addition to the above Maxent framework, the generality of our
theoretical results can be demonstrated by a practical connection
between TEB and another widely used estimator, namely the Lidstone
estimator. We will show that both Frequentist-TEB and Bayesian-TEB
can offer guidance to identify the adaptive rate of probability
correction, which needs to be empirically set in Lidstone.
Accordingly, the so called ``F-Lidstone'' and ``B-Lidstone''
estimators are derived respectively.

Extensive experimental results on a number of synthesized and
real-world datasets demonstrate a promising performance of TEBC
Maxent, F-Lidstone and B-Lidstone, in comparison with various
state-of-the-art density estimation methods.

\section{Related Work}

\label{relworks} The concept of maximum entropy has been existing in
the Machine Learning literature for a long time and has resulted in
various approaches. Its constrained form has been widely applied in
many contexts~\citep{Berger96,Kazama03,Jedynak05}. Recently, there
have been many studies of Maxent with $l_1$-style
regularization~\citep{Khudanpur95,Kazama03,Williams95,Ng04,Goodman04,Krishnapuram05},
$l_2^2$-style regularization~\citep{Lau94,Chen00,Lebanon01,Zhang05}
as well as some other types of regularization such as $l_1 +
l_2^2$-style~\citep{Kazama03}, $l_2$-style
regularization~\citep{Newman77} and a smoothed version of
$l_1$-style regularization~\citep{Dekel03}.~\citet{Altun06} derive
duality and performance guarantees for settings in which the entropy
is replaced by an arbitrary Bregman or Csiszar divergence. A
thoroughly theoretic analysis of regularized Maxent can be found
in~\citep{Miroslav07}.

As another direction to density estimation, there are many smoothing
methods that have been proposed in various contexts, e.g.,
information retrieval tasks~\citep{Zhai01}, speech
recognition~\citep{Chen98} and cryptology~\citep{Good53}. A typical
family of general-purpose smoothing methods is Good-Turing
estimator. All of them use the following equation to calculate the
resulting frequencies of events:
\[
F_X  = \frac{(N_X  + 1)}{T} \cdot \frac{E(N_X  + 1)}{E(N_X )}
\]
where $X$ is an event, $N_X$ is the number of times the event $X$
has been seen, within the sample of size $T$, and $E(n)$ is an
estimate of how many different events that happened exactly for $n$
times. Different variants of Good-Turing estimator, e.g., the
simplest Good-Turing estimator, Simple Good-Turing
estimator~\citep{Gale95} and diminishing-attenuation
estimator~\citep{Orlitsky03}, are based on different calculations of
the $E(\cdot)$.

Another widely used smoothing method is Lidstone
estimator~\citep{Chen98}. Typical variants of Lidstone estimator
include Expected Likelihood Estimator, Laplace estimator and
Add-tiny estimator. From a theoretical point of view, by defining
the attenuation of a probability estimator as the largest possible
ratio between the per-symbol probability assigned to an arbitrary
sized sequence by any distribution and the corresponding probability
assigned by the estimator, it can be shown that the attenuation of
diminishing-attenuation estimators is unity~\citep{Orlitsky03}. Note
that the attenuation analysis is an asymptotic analysis w.r.t. large
sample performance.

For the entropy correction, there are some methods to estimate the
Shannon entropy bias. However, to the best of our knowledge, no
consistent and unbiased correction of Shannon entropy has been
developed.  In principle, the "inconsistency" theorem leads to
several approximations of the Shannon entropy bias
\citep{MillerG,AGCarlton69,Panzeri96,JonathanNC00}. The analytical
approximation of bias given by \citep{Paninski03estimationof} can be
considered more rigorous than predecessors. However, this bias does
not have a closed form and depends on specific prior distribution
and the $c$ \citep{Paninski03estimationof}, and hence it is hard to
be computed in general. Regarding this, Paninski proposed an
estimator, which is consistent even when the $c$ is bounded
(provided that both m and n are sufficiently large)
\citep{TITPaninski04}. However, a general and exact closed-form
formula for Shannon entropy bias is still an open problem.

The rest of this paper is organized as follows:
Section~\ref{theorst} gives a theoretical analysis on two crucial
observations; Section~\ref{TNBC} discusses the estimate of TEBs
(Frequentist-TEB and Bayesian-TEB), introduces TEBC Maxent and
reveals the connection between TEBs and Lidstone estimator;
Section~\ref{eval} gives two model-evaluating criteria; Experiments
on synthesized and real-world datasets are constructed in
Section~\ref{exps} and the experimental results are reported and
discussed. Finally, conclusions and future work are presented in
Section~\ref{ConFut}.

\section{Tsallis Entropy Bias}
\label{theorst}

In this section, we present two theoretical observations, which
motivate and underpin the proposed TEB, TEBC Maxent and TEB-Lidstone
estimator.

\subsection{Notations and Definitions}
\label{def} We use the following notations throughout the rest of
the paper:

$P^{\left( m \right)} \equiv \left\langle {p_1 , \ldots ,p_m }
\right\rangle $ : The underlying real $m$-nomial ($m \ge 2$)
probability distribution, where $\sum\nolimits_{i = 1}^m {p_i } =
1$;

$\widehat P_n^{^{(m)}} \equiv \left\langle {\widehat p_1
,...,\widehat p_m } \right\rangle$ : The sampling distribution of
sample size $n$ w.r.t. $P^{(m)}$, where $\sum\nolimits_{i = 1}^m
{\widehat p_i } = 1$;

$\mathbb {P}^{(m)}$: The set of all possible $m$-nomial ($m \ge 2$)
probability distributions.

$ T_q \left[ {P^{\left( m \right)} } \right] \equiv k\frac{{1 -
\sum\nolimits_{i = 1}^m {p_i^q } }}{{q - 1}} $: The Tsallis entropy
of $P^{\left( m \right)}$ w.r.t. the index $q$, where $k$ is the
Boltzmann constant; We have $ \mathop {\lim }\limits_{q \to 1} T_q
\left[ {P^{\left( m \right)} } \right] = k \cdot H\left[ {P^{\left(
m \right)} } \right]$, where $H\left[ {P^{\left( m \right)} }
\right]$ is the Shannon information entropy of $P^{\left( m
\right)}$.

The Tsallis entropy is the simplest entropy form that extends the
Shannon entropy while maintaining the basic properties but allowing,
if $q \ne 1$, nonextensivity~\citep{Santos97,Abe00}. In this paper,
for the sake of analytical convenience, we always assume $q = 2$ and
neglect the subscript $q$. In addition, without loss of generality,
we omit the Boltzmann constant. It then turns out that $T\left[
{P^{\left( m \right)} } \right] = 1 - \sum\limits_i {p_i^2 }$.

\subsection{Main Results}
\label{mainrst}
\newtheorem{Theorem}{Proposition}
\begin{Theorem}
\label{thrm:one} Given an arbitrary $m$-nomial probability
distribution $P^{(m)}  \in \mathbb{P}^{(m)}$, let $\widehat
P_n^{^{(m)}}$ be the sampling distribution of sample size $n$ with
respect to $P^{(m)}$, $E_{P,n}^{(m)} \left( T \right)$ be the
expected Tsallis entropy of $\widehat P_n^{^{(m)} }$, and $T\left[
{P^{\left( m \right)} } \right] $ be the Tsallis entropy of
$P^{(m)}$, then
\begin{equation}
\label{eq:one} E_{P,n}^{(m)} \left( T \right) = \frac{{n - 1}}
{n}T\left[ {P^{(m)} } \right]
\end{equation}

\begin{proof}
Let $P^{(m)} = \left\langle {p_1,\ldots,p_{m - 1},1 -
\sum\nolimits_{i = 1}^{m - 1} {p_i } } \right\rangle$ be an
$m$-nomial probability distribution. Denote the set of all possible
sampling distributions of sample size $n$ as
\[
\begin{gathered}
  \mathbb{S}_n^{(m)} \equiv \left\{ {\left. {\widehat P_n^{^{(m)} }  \equiv \left\langle {\frac{{x_1 }}
{n}, \ldots ,\frac{{x_{m - 1} }} {n},\frac{{n - \sum\nolimits_{i =
1}^{m - 1} {x_i } }} {n}} \right\rangle } \right|} \right. x_1 ,
\ldots ,x_{m - 1}  \in \mathbb{N}, \ x_1  +  \cdots  + x_{m - 1} \le
n\left.
  \begin{gathered}
   \hfill \\
   \hfill \\
\end{gathered}  \right\} \hfill \\
\end{gathered}
\]
where $x_i$ is the count of the $i^{th}$ nomial. Given $P^{(m)}$,
the occurrence probability of a sampling distribution $\widehat
P_n^{^{(m)} }  \equiv \left\langle {\frac{{x_1 }} {n}, \ldots
,\frac{{x_{m - 1} }} {n},\frac{{n - \sum\nolimits_{i = 1}^{m - 1}
{x_i } }} {n}} \right\rangle  \in \mathbb{S}_n^{(m)}$ is given by
the following equation:
\begin{equation}
\label{eq:two}
\begin{gathered}
  \Pr \left[ {\widehat P_n^{^{(m)} } \left| {P^{(m)} } \right.} \right] = \frac{{n!}}
{{(n - \sum\nolimits_{i = 1}^{m - 1} {x_i } )!\prod\nolimits_{i =
1}^{m - 1} {x_i !} }}\left( {1 - \sum\nolimits_{i = 1}^{m - 1} {p_i } } \right)^{n - \sum\nolimits_{i = 1}^{m - 1} {x_i } } \prod\nolimits_{i = 1}^{m - 1} {p_i^{x_i } }  \hfill \\
\end{gathered}
\end{equation}
Note that we assume $0^0 = 1$ in Formula~\ref{eq:two}. Hence, we
have
\[
\begin{gathered}
  E_{P,n}^{(m)} \left( T \right) = \sum\nolimits_{x_1  = 0}^n {\sum\nolimits_{x_2  = 0}^{n - x_1 }  \cdots  } \sum\nolimits_{x_{m - 1}  = 0}^{n - \sum\nolimits_{i = 1}^{m - 2} {x_i } } {T\left[ {\widehat P_n^{^{(m)} } } \right]} \cdot \Pr \left[ {\widehat P_n^{^{(m)} } \left| {P^{(m)} } \right.} \right] \hfill \\
\end{gathered}
\]
By the definition of Tsallis entropy ($q = 2$), we have
\[
T\left[ {\widehat P_n^{^{(m)} } } \right] \equiv 1 -
\sum\nolimits_{i = 1}^m {\left( {\frac{{x_i }} {n}} \right)} ^2
\]
where we denote $n - \sum\nolimits_{i = 1}^{m - 1} {x_i }$ as $x_m$.
It is convenient to express $E_{P,n}^{(m)} \left( T \right)$ as the
following:
\[
E_{P,n}^{(m)} \left( T \right) = 1 - \sum\nolimits_{i = 1}^m
{E_{P,n}^{(m)} \left( {\frac{{x_i }} {n}} \right)} ^2
\]
where
\[
E_{P,n}^{(m)} \left( {\frac{{x_i }} {n}} \right)^2  \equiv
\sum\nolimits_{x_1  = 0}^n {\sum\nolimits_{x_2  = 0}^{n - x_1 }
\cdots  } \sum\nolimits_{x_{m - 1}  = 0}^{n - \sum\nolimits_{i =
1}^{m - 2} {x_i } } {\left( {\frac{{x_i }} {n}} \right)^2  \cdot \Pr
\left( {\widehat P_n^{^{(m)} } \left| {P^{(m)} } \right.} \right)}
,\;\;i = 1 \ldots m
\]
Note that $E_{P,n}^{(m)} \left( {x_i^2 } \right)$ is just the
moments about the origin of the multinomial distribution and given
by $E_{P,n}^{(m)} \left( {x_i^2 } \right) = n\left( {n - 1}
\right)p_i^2  + np_i$. Hence, we have $E_{P,n}^{(m)} \left(
{\frac{{x_i }} {n}} \right)^2  = \frac{{\left( {n - 1} \right)p_i^2
+ p_i }} {n}$. It turns out that
\[
E_{P,n}^{(m)} \left( T \right) = 1 - \sum\nolimits_{i = 1}^m
{\frac{{\left( {n - 1} \right)p_i^2  + p_i }} {n}}  = \frac{{n - 1}}
{n}\left( {1 - \sum\nolimits_{i = 1}^m {p_i^2 } } \right)
\]
The r.h.s of the last equation is just $T\left[ {P^{\left( m
\right)} } \right]$.
\end{proof}
\end{Theorem}


\newtheorem{Corollary}{Corollary}
\begin{Corollary}
\label{corl:one}
\begin{equation}
\mathop {\lim }\limits_{n \to \infty } E_{P,n}^{(m)} \left( T
\right) = T\left[ {P^{(m)} } \right]
\end{equation}
\begin{proof}
The corollary follows immediately from Formula~\ref{eq:one}.
\end{proof}
\end{Corollary}

The above result is developed in the frequentist framework and hence
corresponds to a Frequentist-TEB. In the following, a uniform
Bayesian TEB (Bayesian-TEB for short) is developed by assuming a
uniform Bayesian prior over all possible $m$-nomial
distributions\footnote{The code package for the numeric evaluation
of Proposition 1 and 2 is provided at \\
\url{http://www.comp.rgu.ac.uk/staff/pz/TEBC/Proposition_Validation_Codes.zip}}.


\begin{Theorem}
\label{thrm:two} Given the uniform probability metric over
$\mathbb{P}^{(m)}$, the expectation of $E_{P,n}^{(m)}$, i.e.
$E_n^{(m)}$, is given by
\begin{equation}
\label{thrm:two-one} E_n^{(m)} \left( T \right) = \frac{{(n - 1)
\cdot (m - 1)}}{{n \cdot (m + 1)}}
\end{equation}

\begin{proof}
By the definition of mathematical expectation, we have
\[
\begin{gathered}
 E_n^{(m)} \left( T \right) =
 \frac{1}{{Z^{(m - 1)} }}\int_0^1 {\int_0^{1 - p_1 } { \cdots \int_0^{1 - \sum\nolimits_{i = 1}^{m - 2} {p_i } } {E_{P,n}^{(m)} \left( T \right)}}} ~dp_{m - 1} dp_{m - 2}  \cdots dp_1 \hfill  \\
\end{gathered}
\]
where $Z^{(m - 1)}$ is the normalization factor determined by the
$(m-1)$-order integral operator,
\[
\begin{gathered}
  Z^{(m - 1)}  = \int_0^1 {\int_0^{1 - p_1 } { \cdots \int_0^{1 - \sum\nolimits_{i = 1}^{m - 2} {p_i } } {dp_{m - 1} dp_{m - 2}  \cdots dp_1 } } }  \hfill \\
  \quad \quad \quad \ = \frac{1}{{(m - 1)!}} \hfill \\
\end{gathered}
\]
Expands and rewrites $E_{P,n}^{(m)} \left( T \right)$ as the
following:
\begin{equation}
\label{thrm:two-two}
\begin{gathered}
  E_{P,n}^{(m)} \left( T \right) = \frac{{2\left( {n - 1} \right)}}{n}\left( {\sum\nolimits_{i = 1}^{m - 1} {p_i } } {- \sum\nolimits_{i = 1}^{m - 1} {p_i^2 }  - \sum\nolimits_{i = 1}^{m - 1} {\sum\nolimits_{j = i + 1}^{m - 1} {p_i  \cdot p_j } } } \right) \hfill \\
\end{gathered}
\end{equation}
Hence, we have
\[
\begin{gathered}
  E_n^{(m)} \left( T \right) = \frac{{2\left( {n - 1} \right)}}{{Z^{(m - 1)}  \cdot n}} \cdot \int_0^1 {\int_0^{1 - p_1 } { \cdots \int_0^{1 - \sum\nolimits_{i = 1}^{m - 2} {p_i } } }} \hfill \\
  ~~~~~~~~~~~(\sum\nolimits_{i = 1}^{m - 1} {p_i}  - \sum\nolimits_{i = 1}^{m - 1} {p_i^2 }  - \sum\nolimits_{i = 1}^{m - 1} {\sum\nolimits_{j = i + 1}^{m - 1} {p_i  \cdot p_j } } ) \hfill \\
  ~~~~~~~~~~~dp_{m - 1} dp_{m - 2}  \cdots dp_1 \hfill
\end{gathered}
\]

To simplify the notations, we define the $(m-1)$-order integral
operator:
\begin{equation}
\label{thrm:two-three} L^{(m - 1)}  \equiv \int_0^1 {\int_0^{1 - p_1
} { \cdots \int_0^{1 - \sum\nolimits_{i = 1}^{m - 2} {p_i } } {dp_{m
- 1} dp_{m - 2} \cdots dp_1 } } }
\end{equation}
We denote $L^{(m - 1)} \left( {f\left( {p_1 , \ldots ,p_{m - 1} }
\right)} \right)$ as
\[
\begin{gathered}
 L^{(m - 1)} \left( {f\left( {p_1 , \ldots ,p_{m - 1} } \right)} \right)
 \equiv \int_0^1 {\int_0^{1 - p_1 } { \cdots \int_0^{1 - \sum\nolimits_{i = 1}^{m - 2} {p_i } } {} } } f\left( {p_1 , \ldots ,p_{m - 1} } \right)dp_{m - 1} dp_{m - 2}  \cdots dp_1 \hfill \\
\end{gathered}
\]
Then the proof of Formula~\ref{thrm:two-one} is reduced to solve the
closed-form of
\[
\begin{gathered}
L^{m-1}  (\sum\nolimits_{i = 1}^{m - 1}
{p_i } - \sum\nolimits_{i = 1}^{m - 1} {p_i^2 } - \sum\nolimits_{i = 1}^{m - 1} {\sum\nolimits_{j = i + 1}^{m - 1} {p_i  \cdot p_j } }) \\
\end{gathered}
\]
Due to the symmetry of integral domain, $L^{(m - 1)}$ has the
following properties:
\begin{enumerate}
\item[(a)] $L^{(m - 1)} (p_i ) = L^{(m - 1)} (p_j ),1 \le i,j \le m -
1$
\item[(b)] $L^{(m - 1)} (p_i^2 ) = L^{(m - 1)} (p_j^2 ),1 \le i,j \le m -
1$
\item[(c)] $L^{(m - 1)} (p_i  \cdot p_j ) = L^{(m - 1)} (p_k  \cdot p_l ),1 \le
i,j \le m - 1,i \ne j,k \ne l$
\end{enumerate}
Therefore, if the general term formulae of $L^{(m - 1)} (p_i )$,
$L^{(m - 1)} \left( {p_i^2 } \right)$ and $L^{(m - 1)} \left( {p_i
\cdot p_j } \right), i\ne j$, and their term numbers are available,
the general term formula of $E_n^{(m)} \left( T \right)$ could be
obtained directly.

The following general term formulae can be verified:
\begin{enumerate}
\item[(a')] $L^{(m - 1)} (p_i ) = \frac{1}{{m!}}$
\item[(b')] $L^{(m - 1)} \left( {p_i^2 } \right) = \frac{2}{{\left( {m + 1}
\right)!}}$
\item[(c')] $L^{(m - 1)} \left( {p_i  \cdot p_j } \right) = \frac{1}{{\left( {m +
1} \right)!}}$
\end{enumerate}
then
\[
\begin{gathered}
 E_n^{(m)} \left( T \right) = \frac{{2\left( {n - 1} \right)}}{{Z^{(m - 1)}  \cdot n}} \cdot L^{(m - 1)} (\sum\nolimits_{i = 1}^{m - 1} {p_i }  - \sum\nolimits_{i = 1}^{m - 1} {p_i^2 } - \sum\nolimits_{i = 1}^{m - 1} {\sum\nolimits_{j = i + 1}^{m - 1} {p_i  \cdot p_j } } ) \hfill \\
 \quad \quad \quad \ \ \ = \frac{{2\left( {n - 1} \right)}}{{Z^{(m - 1)}  \cdot n}}\left[ {\frac{{m - 1}}{{m!}} - \frac{{2\left( {m - 1} \right)}}{{\left( {m + 1} \right)!}} - \frac{{\left( {m - 1} \right) \cdot \left( {m - 2} \right)}}{{2\left( {m + 1} \right)!}}} \right] \hfill \\
\end{gathered}
\]
Recall that $Z^{(m - 1)}  = \frac{1}{{(m - 1)!}}$, and after some
simplification steps, Formula~\ref{thrm:two-one} is obtained from
the above equation, which completes the proof of
Proposition~\ref{thrm:two}.
\end{proof}
\end{Theorem}

\begin{Corollary}
\label{corl:two} Let $E^{(m)} \left( T \right) = \frac{1} {{Z^{(m -
1)} }}\int_0^1 {\int_0^{1 - p_1 } { \cdots \int_0^{1 -
\sum\nolimits_{i = 1}^{m - 2} {p_i } } {T\left[ {P^{\left( m
\right)} } \right]dp_{m - 1} dp_{m - 2}  \cdots dp_1 } } }$, where
$Z^{(m - 1)}$ is the normalization factor $\frac{1} {{(m - 1)!}}$.
Then we have
\begin{equation}
\label{eq:corl2} \mathop {\lim }\limits_{n \to \infty } E_n^{(m)}
\left( T \right) = E^{(m)} \left( T \right)
\end{equation}
\begin{proof}
The corollary follows directly from the fact that $E^{(m)} \left( T
\right)$ can be given by the r.h.s of Formula~\ref{thrm:two-two},
except for a multiplicative factor $\frac{{n - 1}} {n}$.
\end{proof}
\end{Corollary}

\section{Density Estimate based on Tsallis Entropy Bias} \label{TNBC}

Based on the above theoretic results, we first discuss the issue on
the estimate of Tsallis entropy bias, and then propose density
estimate methods in Maxent framework and Lidstone framework,
respectively.

\subsection{On Estimation of Tsallis Entropy Bias}
\label{estimation} To apply the result of
Proposition~\ref{thrm:one}, the frequentist Tsallis entropy bias
(Frequentist-TEB) should be effectively estimated so that the
Tsallis entropy of the sampling distribution $\widehat P_n^{^{(m)}}$
can be compensated accordingly. According to Formula~\ref{eq:one},
the expected Tsallis entropy of $\widehat P_n^{^{(m)}}$, i.e.,
$E_{P,n}^{(m)} \left( T \right)$, is ${{(n - 1)} \mathord{\left/
{\vphantom {{(n - 1)} n}} \right. \kern-\nulldelimiterspace} n}$ of
the Tsallis entropy of the underlying real distribution $P^{(m)}$,
i.e., $T\left[ {P^{(m)} } \right]$. We denote the estimation of
$E_{P,n}^{(m)} \left( T \right)$ as $\widehat E_{P,n}^{(m)} \left( T
\right)$, then $\frac{n}{{n - 1}}\widehat E_{P,n}^{(m)} \left( T
\right)$ can be considered as an estimation of $T\left[ {P^{(m)} }
\right]$. Hence, the estimation of frequentist Tsallis entropy bias,
i.e. Frequentist-TEB, is given by£º
\begin{equation}
\label{eq:six} \Delta T = \frac{n} {{n - 1}}\widehat E_{P,n}^{(m)}
\left( T \right) - T\left[ {\widehat P_n^{^{(m)} } } \right]
\end{equation}
The simplest (and unbiased) estimation of $\widehat E_{P,n}^{(m)}
\left( T \right)$ is given by $T\left[ {\widehat P_n^{^{(m)} } }
\right]$, and hence the corresponding estimated TEB is given by
\begin{equation}
\label{eq:seven} \Delta T = \frac{1} {{n - 1}}T\left[ {\widehat
P_n^{^{(m)} } } \right]
\end{equation}

\newtheorem{Remark}{Remark}

\begin{Remark}
\label{Rem:one} $T\left[\widehat P_n^{^{(m)}}\right]$ is an unbiased
estimate of $E_{P,n}^{(m)}\left( T \right)$. Therefore,
Formula~\ref{eq:seven} gives an unbiased correction of
$T\left[P^{^{(m)}}\right]$. In addition,  the consistency of this
correction follows from the law of large numbers  (if $m$ is finite)
or central limit theorem for multinomial sums \citep{MorrisAS75} (if
$m$ is infinite).

\end{Remark}

Surprisingly, experimental results (detailed in Section \ref{exps})
show the TEBC Maxent and TEB-Lidstone estimator based on this naive
estimator can outperform all comparative density estimation models
in most cases.

In many cases, using statistical re-sampling techniques, e.g.,
Bootstrap~\citep{Larry06}, we can achieve more accurate estimations
of $T\left[ {P^{(m)} } \right]$, and hence obtain better estimations
of Frequentist-TEB. In the following, we give an estimation
procedure of $T\left[ {P^{(m)} } \right]$, which is optimal in the
sense of the least squared error.

Let us rewrite Formula~\ref{eq:one} so that $E_{P,n}^{(m)} \left( T
\right)$ is expressed by a function of $n$
\[
E_{P,n}^{(m)} \left( T \right) = K \cdot \frac{{n - 1}} {n}
\]
where $K$ is a constant slope and determined by $T\left[ {P^{(m)} }
\right]$. We can estimate $E_{P,i}^{(m)} \left( T \right),\;1 \le i
\le n$ by re-sampling techniques and obtain $\widehat E_{P,i}^{(m)}
\left( T \right),\;1 \le i \le n$. Note that the re-sampling is
meaningful only if $i < n$. The remaining task is to solve an
unconstrained quadratic program so that the squared error
\begin{equation}
\label{eq:7.5}\sum\nolimits_{i = 1}^n {\left[ {\widehat
E_{P,i}^{(m)} \left( T \right) - \frac{{i - 1}} {i}K} \right]} ^2
\end{equation}
is minimized. This minimum corresponds to the zero point of the
first derivative in the cost function
\[
0 = \frac{\partial } {{\partial K}}\sum\nolimits_{i = 1}^n {\left[
{\widehat E_{P,i}^{(m)} \left( T \right) - \frac{{i - 1}} {i}K}
\right]} ^2
\]
By expanding the above equation, it turns out that the estimated
slope is given by
\[
\widehat K = \frac{{\sum\nolimits_{i = 1}^n {(i - 1)/i \cdot
\widehat E_{P,i}^{(m)} \left( T \right)} }}{{\sum\nolimits_{i = 1}^n
{(i - 1)^2 /i^2 } }}
\]
where $\widehat K$ is the estimation of $T\left[ {P^{(m)} }
\right]$. Note that, in practice, it is often sufficient to only
involve an appropriate subset of $\widehat E_{P,i}^{(m)} \left( T
\right),\;1 \le i \le n$, in the cost function~(\ref{eq:7.5}) to
construct the estimate of $T\left[ {P^{(m)} } \right]$.

However, even with the re-sampling technique, in the case that the
sampling is seriously inadequate, it seems still difficult to
estimate $E_{P,n}^{(m)} \left( T \right)$ accurately, which might in
turn result in inaccurate Frequentist-TEB. In this case, some
Bayesian prior over the space of all possible real distributions
might be more useful. The results of Proposition~\ref{thrm:two} and
Corollary~\ref{corl:two} guide the construction of uniform
Bayesian-TEB, i.e.,
\begin{equation}
\label{UB-TEB} \Delta T = \frac{{m - 1}}{{n \cdot \left( {m + 1}
\right)}}
\end{equation}
by assuming the uniform probability metric over all possible real
distributions. Note that uniform Bayesian-TEB is directly obtained
by computing the difference between the expected Tsallis entropy of
all possible real distributions and the expectation of
$E_{P,n}^{(m)} \left( T \right)$, i.e. $E_{n}^{(m)} \left( T
\right)$, w.r.t. uniform prior. Hence, it avoids the estimation of
$E_{P,n}^{(m)} \left( T \right)$.

\begin{Remark}
\label{Rem:two} Corollary \ref{corl:two} shows that $E_n^{(m)}\left(
T \right)$ is the asymptotically unbiased estimate of $E^{(m)}\left(
T \right)$ and can be unbiased if the Bayesian-TEB
$\frac{m-1}{n(m+1)}$ is compensated. In addition, this corrected
estimator is also consistent with $E^{(m)}\left( T \right)$.
\end{Remark}

In summary, Remark1 and Remark2 show that the estimates of the
Tsallis entropy bias, w.r.t both Frequentist and Bayesian
frameworks, can be considered sound in the sense of unbiasedness and
consistency. Note that the Shannon entropy estimate lacks these
guarantees.

\subsection{TEBC Maxents}
\label{model} In this subsection, we propose three Tsallis entropy
bias compensation (TEBC) Maxents to compute resulting distributions.
These distributions are most similar to the sampling distribution
w.r.t. different similarity criteria, subject to the constraint that
the estimated Frequentist/Bayesian-TEB, denoted as $\Delta T$, is
forcibly compensated. This strategy can help to alleviate the
overfitting and underfitting problems.

Given any $m$-nomial sampling distribution $\widehat P_n^{^{(m)} }
\equiv \left\langle {\widehat p_1 , \ldots ,\widehat p_m
}\right\rangle$ of sample size $n$ and the estimated $\Delta T$, the
TEBC Maxents can be constructed to compute the resulting
distribution ${\bar P}^{\left( m \right)} \equiv \left\langle {{\bar
p}_1 , \ldots ,{\bar p}_m } \right\rangle $ w.r.t. the criterion of
$l_2^2$ norm, Jensen-Shannon (JS) divergence (see
Formula~\ref{eq:ten} for details) or Maximum Likelihood,
respectively:

{\bf Model 1: $l_2^2$ Tsallis Entropy Bias Compensation
($l_2^2$-TEBC)}
\begin{equation}
\label{model1}
\begin{gathered}
  \mathop {\min }\limits_{\bar P^{(m)} }~~\sum\nolimits_{i = 1}^m {\left( {\bar{p}_i  - \widehat p_i } \right)^2 }  \hfill \\
  s.t.\quad T\left[ {\bar P^{(m)} } \right] \ge T\left[ {\widehat P_n^{(m)} } \right]{\text{ + }}\Delta T \hfill \\
  Certain\verb# # Constraints
\end{gathered}
\end{equation}

{\bf Model 2: JS-Divergence Tsallis Entropy Bias Compensation
(JSD-TEBC)}
\begin{equation}
\label{model2}
\begin{gathered}
 \mathop {\min }\limits_{\bar P^{(m)} }~~JSD\left[ {\bar{P}^{(m)} \left | {\widehat P_n^{(m)} } \right.} \right] \hfill\\
 s.t.\quad T\left[ {\bar P^{(m)} } \right] \ge T\left[ {\widehat P_n^{(m)} } \right]{\text{ + }}\Delta T \hfill\\
 Certain\verb# # Constraints
\end{gathered}
\end{equation}
where $JSD\left[ { \cdot | \cdot } \right]$ denotes the
JS-divergence.

Note that a common statistic to measure the divergence between two
probability distributions is the Kullback-Leibler (KL) divergence.
Despite of the computational and theoretical advantages of
KL-divergence, it is not symmetric in its arguments. Reversing the
arguments in the KL-divergence function can yield substantially
different results. Furthermore, $KL(P,Q)$ may be seriously
underestimated if $P$ involves zero terms since
$\lim_{p_i\to0}p_i\log\frac{{p_i}}{{q_i}}=0$. Besides, $KL(P,Q)$ is
sensitive to penalty terms used in the case of $q_i = 0$. Hence, we
apply a symmetrized variant of KL-divergence, i.e., JS-divergence,
instead.
\begin{equation}
\label{eq:ten} JSD\left[ {\left. P \right |Q} \right] \equiv
\frac{1} {2}D\left[ {\left. P \right |M} \right] + \frac{1}
{2}D\left[ {\left. Q \right |M} \right]
\end{equation}
where $D\left[ {P\left | M \right.} \right]$ is the KL-divergence
from $P$ to $M$ and $M = (P + Q) / 2$.

{\bf Model 3: Maximum Likelihood Tsallis Entropy Bias Compensation
(ML-TEBC)}
\begin{equation}
\label{model3}
\begin{gathered}
 \mathop {\max \limits_{\bar P^{(m)} }~~\log \left\{ {\Pr \left[ {\widehat P_n^{(m)} \left| {\bar{P}^{(m)} } \right.} \right]} \right\}} \hfill\\
 s.t.\quad T\left[ {\bar P^{(m)} } \right] \ge T\left[ {\widehat P_n^{(m)} } \right]{\text{ + }}\Delta T \hfill\\
 Certain\verb# # Constraints
\end{gathered}
\end{equation}
where $\Pr \left[ {\widehat P_n^{(m)} \left| {\bar P^{(m)} }
\right.} \right]$ is given by Formula~\ref{eq:two}.

In Models 1, 2 and 3, all objective functions aim at forcing the
resulting distribution similar to sampling distribution as well as
possible. This is to counter the underfitting problem. Meanwhile,
the TEB constraint is used to increase the entropy of the resulting
distribution and then compensate the Tsallis entropy bias, in order
to make the resulting distribution approximate the underlying real
distribution. This is to avoid the overfitting problem. From another
point of view, in the expected sense, the objective function aims at
reducing the Tsallis entropy, while the TEB constraint is adopted to
necessarily increase the Tsallis entropy. Through this joint effort,
the objective function forces the TEB constraint to hold as
equality, which is consistent with our previous theoretical
analysis.

In implementation, all the above objective functions and constraints
are convex. Therefore, TEBC Maxents can be globally solved by
efficient methods, e.g., the interior method ~\citep{Stephen04}. In
specific application contexts, TEBC Maxents should include certain
constraints, which are derived from reliable prior information and
do not involve empirical threshold parameters. A specific example on
the form of certain constraints is given in our experiment (see
Section~\ref{Constraints} for details).


TEBC Maxents can be constructed with Frequentist-TEB or
Bayesian-TEB. In the cases that sampling process is seriously
inadequate, the Bayesian TEBC Maxents are expected to have stable
performance. The cause is that, given uniform Bayesian prior and an
inadequate sampling, e.g., $n \approx m$, the standard deviation of
$E_{P,n}^{(m)}(T)$ tends to be negligible compared to uniform
Bayesian-TEB, which implies a relatively stable estimation of
uniform Bayesian-TEB. The detailed proof is given in
Proposition~\ref{thrm:four} of Appendix A.

\subsection{TEB-Lidstone Estimators}
\label{connection} There is a natural connection between TEBs and
Lidstone estimator. Lidstone's law of succession suggests the family
of Lidstone estimators in the following form:
\[
\bar p_i  = \frac{{x_i  + f}}{{n + f \cdot m}}
\]
where $n$ is the sample size, $m$ is the number of nomials, $x_i$ is
the count of the $i^{th}$ nomial and $f$ is a parameter indicating
the rate of probability correction (normally between 0 and 1). When
$f = 0.5$, it turns out to be the well-known Expected Likelihood
Estimator (ELE), i.e.,
\[
\bar p_i  = \frac{{x_i  + 0.5}}{{n + 0.5 \cdot m}}
\]
Another two common Lidstone estimators are add-one estimator ($f =
1$) and add-tiny estimator ($f = 1/n$). The smaller $f$ is, the less
probability mass it compensates for underestimations. There exist
some explanations on the selection of parameter $f$. For example,
ELE gives a Bayesian justification by assuming a uniform prior for a
binomially distributed variable~\citep{Box73}. However, in general
cases, $f$ is empirically configured.

TEBs offer a set of criteria, either of which analytically
identifies an adaptive $f$ w.r.t. a specific input sample, and
derives the TEB-Lidstone estimator. The fundamental idea is to solve
such an $f$ so that the Tsallis entropy bias of the input sample is
quantitatively compensated. That is
\[
1 - \sum\nolimits_i {\left( {\frac{{x_i  + f}}{{n + f \cdot m}}}
\right)} ^2  = T\left[ {\widehat P_n^{\left( m \right)} } \right] +
\Delta T
\]
where $\Delta T$ can be Frequentist-TEB or Bayesian-TEB. Let $\alpha
\equiv 1 - \left( {T\left[ {\widehat P_n^{\left( m \right)} }
\right] + \Delta T} \right)$. It turns out that we have the
following quadratic equation in the single variable $f$:
\begin{equation}
\label{solveF} \left( {\alpha m^2  - m} \right)f^2  + 2n\left(
{\alpha m - 1} \right)f + \left( {kn^2  - \sum\nolimits_{i = 1}^m
{x_i^2 } } \right) = 0
\end{equation}

Occasionally, Formula~\ref{solveF} has not real roots. In this case,
we can simply select $f$ corresponding to the minimum (if $\alpha
m^2 - m > 0$) or the maximum (if $\alpha m^2  - m < 0$) of the l.h.s
of Formula~\ref{solveF}. Consequently, the so called ``F-Lidstone''
and ``B-Lidstone'' estimators are derived on Frequentist-TEB and
Bayesian-TEB, respectively.

Note that, in principle, the Tsallis entropy bias can also serve to
identify the parameters of some other estimators, e.g., the
multiplicative parameter of Good-Turing estimator. We omit the
computation details here.

\section{Evaluation Criteria} \label{eval}

The performance of a density estimation method can be directly
evaluated by measuring the similarity between its solution and the
underlying real distribution. As mentioned in Formula~\ref{eq:ten},
JS-divergence can be considered as a candidate criterion. In
addition, we use the expected log loss (the expect negative
normalized log likelihood~\citep{Miroslav07}) as another similarity
criterion. The log loss of a resulting distribution $\bar P^{(m)}
\equiv \left\langle {\bar p_1 , \ldots ,\bar p_m } \right\rangle$
with respect to the sample ${\mathbf{x}} \equiv \left\langle {x_1
,x_2 , \ldots ,x_m } \right\rangle$ is defined as:
\begin{equation}
\label{eq:eleven} L_{\bar P^{(m)}} \left( {\bf{x}} \right) \equiv -
\log \bar p_1^{x_1 } \bar p_2^{x_2 }  \cdots \bar p_m^{x_m } = -
\sum\limits_{i = 1}^m {x_i \log \bar p_i }
\end{equation}
Recall that, given a underlying real $m$-nomial distribution
$P^{\left( m \right)}  \equiv \left\langle {p_1 , \ldots ,p_m }
\right\rangle$ , the occurrence probability of a sample
${\mathbf{x}} \equiv \left\langle {x_1 ,x_2 , \ldots ,x_m }
\right\rangle$ of size $n$ can be expressed by
\begin{equation}
\label{eq:twelve} \Pr \left[ {{\mathbf{x}}\left| {P^{\left( m
\right)} } \right.} \right] = \frac{{n!}} {{x_1 !x_2 ! \cdots x_m
!}} \cdot p_1^{x_1 } p_2^{x_2 }  \cdots p_m^{x_m }
\end{equation}
Hence, we can define the expected log loss of $\bar P^{(m)}$ w.r.t.
$P^{(m)}$ as
\begin{equation}
\label{eq:thirteen} E_L \left[ {\bar P^{(m)} \left| {P^{(m)} }
\right.} \right] \equiv \sum\nolimits_{{\mathbf{x}} \in X} {L_{\bar
P^{(m)} } \left( {\mathbf{x}} \right)\Pr \left[ {{\mathbf{x}}\left|
{P^{(m)} } \right.} \right]}
\end{equation}
where $X$ stands for the set of all possible samples of size $n$. By
substituting the r.h.s of Formula~\ref{eq:thirteen} by Formula
\ref{eq:eleven} and \ref{eq:twelve}, it can be checked that

\begin{equation}
\label{eq:fourteen}
\begin{array}{l}
  E_L \left[ {\bar P^{(m)} \left| {P^{(m)} } \right.} \right] = \sum\limits_{{\bf x} \in X} {\sum\limits_{i = 1}^m { - x_i \log \bar p_i }  \cdot \frac{{n!}}{{x_1 !x_2 ! \cdots x_m !}}p_1^{x_1 } p_2^{x_2 }  \cdots p_m^{x_m } }  \\
  \quad \quad \quad \quad \quad \quad \quad \quad =  - \sum\limits_{i = 1}^m {\log \bar p_i \sum\limits_{{\bf x} \in X} {x_i \frac{{n!}}{{x_1 !x_2 ! \cdots x_m !}}p_1^{x_1 } p_2^{x_2 }  \cdots p_m^{x_m } } }  \\
  \quad \quad \quad \quad \quad \quad \quad \quad =  - \sum\limits_{i = 1}^m {\log \bar p_i } \left( {np_i } \right) \\
  \quad \quad \quad \quad \quad \quad \quad \quad =  - n\sum\limits_{i = 1}^m {p_i \log \bar p_i }  \\
 \end{array}
\end{equation}

To more systematically measure the performance of different
algorithms w.r.t a specific evaluation criterion, we introduce the
{\bf Performance Score (PS)} as below:

\begin{equation}\label{eq:ps}
PS_D^C (A) =  {\frac{{C_D(A) - C_D(Worst)}}{{C_D(Best) -
C_D(Worst)}}}
\end{equation}
where $A$ stands for an algorithm under evaluation, $C$ represents
an evaluation criterion and $D$ denotes a specific dataset; the
performance value $C_D(A)$ is evaluated by criterion $C$ for
algorithm $A$ running on dataset $D$, and $C_D(Best)$ and
$C_D(Worst)$ are the performance values of the best and worst
algorithms on $D$, respectively.

\section{Experiments}
\label{exps} In this section, we construct three sets of
experiments\footnote{The source code is available at  \\
\url{http://www.comp.rgu.ac.uk/staff/pz/TEBC/TEB_Experiment_Codes.zip}}.
First, if reliable prior information is available, in Maxent
framework, TEBC Maxents' performance will be evaluated in comparison
with standard Maxent, GME (a $l_1$-regularized Maxent which has PAC
guarantee of performance)~\citep{Miroslav07}, as well as Maxents
based on Shannon entropy bias (SEB)~\citep{MillerG} which is
described in Section~\ref{Maxents}. Second, in case that reliable
prior information is not available, we will verify the effectiveness
of TEB-Lidstone, by comparing their performance with comparative
Lidstone and Good-Turing estimators, together with SEB-Lidstone
which is derived from the above SEB in a similar way to TEB-Lidstone
(described in Section~\ref{Lidstones}). In all the above
experimental settings, both synthesized and real-world datasets are
employed. For TEBs, Bayesian-TEB is calculated by
Formula~\ref{UB-TEB}, and Frequentist-TEB is estimated by the naive
estimator given in Formula~\ref{eq:seven}, which is more efficient
in large-scale experiments and also can give a satisfying
performance in both Maxent and Lidstone frameworks.


\subsection{Datasets Description}
First, synthesized probability distributions are generated to serve
as the underlying real $m$-nomial distributions $P^{(m)}$. To this
end, we adopt a simple Monte Carlo method to randomly draw $m$
positive points from a source distribution, and then normalize them
to form an underlying real distribution. The source distributions we
used include uniform distribution $U(0, 1)$, the absolute value of
standard normal distribution $|N(0, 1^2 )|$, normal distribution
$N(3, 1^2 )$, ${\chi}^2$ distribution ${\chi}^2(10)$, binomial
distribution $B(30, 0.2)$ and beta distribution $\beta(3, 6)$. After
the underlying real distribution is generated, a sample of size $n$
is drawn from it, which can be then used to calculate the sampling
distribution $\widehat P^{(m)}$.

We also adopt four real-world datasets:
UCI-Dexter\footnote{http://archive.ics.uci.edu/ml/machine-learning-databases/dexter/DEXTER/},
UCI-Statlog\footnote{http://archive.ics.uci.edu/ml/machine-learning-databases/statlog/satimage/},
UCI-ISOLET\footnote{http://archive.ics.uci.edu/ml/machine-learning-databases/isolet/}
and
UCI-Sonar\footnote{http://archive.ics.uci.edu/ml/machine-learning-databases/undocumented/connectionist-bench/sonar/},
in order to generate the underlying real distributions

{\bf Text dataset: Dexter}

UCI-Dexter is a text dataset, containing 2000+300 documents. Each
document is represented as a 20000-term count vector. Before the
text dataset is actually employed, it is preprocessed by dropping
the terms that occur too frequently, i.e., the ``stop words''. After
the preprocessing step, $m$ terms could be randomly selected from
the whole term set and the frequencies of these selected terms are
considered as the underlying real distribution $P(m)$. Then, we
randomly choose a bag (size $n$) of words from all the documents as
a sample. The frequencies of these terms in the sample are
calculated and considered as the sampling distribution $\widehat
P(m)$.

{\bf Non-text datasets: Statlog, ISOLET, Sonar}

UCI-Statlog (Landsat Satellite) dataset consists of all possible
$3\times3$ neighborhoods in a 82$\times$100 pixel sub-area of a
single scene which is represented by four digital images in
different spectral bands. A sample is then defined as the pixel
values of each $3\times3$ neighborhood in the four spectral bands
(hence 4$\times$9=36 features in total). The size of the dataset is
4435+2000.

UCI-ISOLET dataset includes 150 subjects speaking the name of each
letter of the alphabet twice. The speakers are divided into groups
of 30 speakers. There are 617 real-value features including spectral
coefficients, contour features, sonorant features, pre-sonorant
features, and post-sonorant features but in an unknown order.

UCI-Sonar contains 111 patterns obtained by bouncing sonar signals
off a metal cylinder at various angles and under various conditions,
and 97 patterns obtained from rocks under similar conditions. The
transmitted sonar signal is a frequency-modulated chirp, rising in
frequency. The data set contains signals obtained from a variety of
different aspect angles, spanning 90 degrees for the cylinder and
180 degrees for the rock. Each pattern is a set of 60 numbers in the
range 0.0 to 1.0. Each number represents the energy within a
particular frequency band, integrated over a certain period of time.

For the above three non-text datasets, in order to generate the
$m$-nomial underlying real distribution, we simply partition a
randomly selected feature into $m$ intervals covering the whole
range of this single feature. Then the number of instances in each
interval is counted and finally the underlying real distribution
$P^{(m)}$ is formed by normalizing the count vector. The sampling
process is to first randomly choose $n$ instances and distribute
them into the corresponding intervals based on their feature value,
and then form the sampling distribution $\widehat P^{(m)}$ by
normalizing this sampling count vector.

\subsection{TEBC Maxents vs. Comparative Maxents}
\label{priorExp} Maxent is widely used due to its effective use of
reliable prior information. In this set of experiments where the
reliable prior information is given, TEBC Maxents and other Maxents
are compared in terms of their density estimation performance.

\subsubsection{Maxents}
\label{Maxents}

Various forms of Maxents are tested, including Frequentist TEBC
Maxents (F-$l^2_2$-TEBC, F-ML-TEBC and F-JSD-TEBC), Bayesian TEBC
Maxents (B-$l^2_2$-TEBC, B-ML-TEBC and B-JSD-TEBC), standard Maxent
(SME for short), and GME \citep{Miroslav07} (implemented by
$l_1$-SUMMET and $l_1$-PLUMMET, which stand for selective-update and
parallel-update algorithms for $l_1$-regularization Maxent,
respectively). In addition, we also construct three Maxents based on
Shannon entropy bias (SEB)~\citep{MillerG}. SEB Maxents are similar
to TEBC Maxent (Model 1-3) except that the TEB constraint is
replaced by the following SEB constraint:
\begin{equation}
\label{SEB} S\left[ {\bar P^{(m)} } \right] \ge S\left[ {\widehat
P_n^{(m)} } \right]{\text{ + }}\Delta S
\end{equation}
where $S[\cdot]$ denotes the Shannon entropy of some probability
distribution and $\Delta S = (m-1)/(2n)$ denotes SEB\footnote{In the
original formula, an estimate of $m$ is used instead of the real
one. In our context, $m$ is known in advance and hence the
estimation can be avoided.}. We use $\frac{m-1}{2n}$ as the
correction since it is simple in form and frequently-used. Note that
it can not be considered an unbiased correction, in a strict
sense~\citep{Paninski03estimationof}. By substituting the SEB
constraint for the TEB constraint in Model 1-3, three new Maxents
are introduced in the experiment, namely $l^2_2$-SEB, ML-SEB and
JSD-SEB.

Note that in the following, we adopt ``Sample'' to represent the
method using the sampling distribution as the resulting distribution
directly.

\subsubsection{Certain and Uncertain Constraints}
\label{Constraints}

Two kinds of constraints are involved. One is certain constraints,
and the other is uncertain constraints. Certain constraints are
derived from reliable prior information, which is incomplete
information of the underlying real distribution. Specifically,
certain constraints can be represented by a set of constraints as
follows:

\begin{equation}\label{eq:cercons}
\sum\limits_{i \in S_c} {\bar p_i}  = a_c^* =  \sum\limits_{i \in
S_c} {p_i } \ \ \ \ \forall c \in C
\end{equation}
where $S_c$ is a subset of $\{1,2,\ldots,m\}$. In order to form this
subset, we randomly choose a number $|S_c|$ from
$\{1,2,\ldots,m-1\}$, and then randomly select $|S_c|$ indexes from
$\{1,2,\ldots,m\}$. If we do this step $k$ times, $k$ certain
constraints can be derived, and then the set $C$ of all certain
constraints is formed.

Uncertain constraints are derived from sampling information,
together with empirical threshold parameters to control the
similarity between the resulting distribution and the sampling
distribution. Specifically, for every $i$, if $\widehat p_i \ge th$,
then we construct an uncertain constraint represented by {\bf Box
Constraint}:

\begin{equation}\label{eq:uncercons}
\left| {\bar p_i - \widehat p_i} \right| \le \delta
\end{equation}
where $\delta$ and $th$ are threshold parameters. In our
experiments, we fix $th$ as $0.2/m$, and adjust $\delta$ to find
relatively optimal performance for standard Maxent and GME.

\subsubsection{Parametric Configuration}
\label{params}

For all the models, the same parameters are used, including
generating times, bin number, sample size, and sampling times.
Generating times is the number of times to generate the underlying
real distribution. Sampling times is the number of times to draw the
sampling distribution from the given underlying real distribution.
Note that in order to avoid the zero probability of any bin in the
$m$-nomial underlying real distribution, the bin number $m$ should
be set properly for each real-world dataset in terms of its scale.
For example, Sonar dataset has a relatively small number of data
points. Therefore, we let $m_{Sonar} = 30$ to avoid a degenerated
$m$-nomial underlying real distribution.

All Maxents involve the same certain constraints. TEBC Maxents and
SEB Maxents adopt the corresponding TEB and SEB constraints, while
other three Maxents use uncertain constraints. For uncertain
constraints, we choose $\delta$ in Formula~\ref{eq:uncercons} from
$[1e-4, 1.6e-3]$ with increment $1e-4$. The performance of TEBC
Maxents will be compared with the performance of comparative Maxents
with the optimal $\delta$. Under this optimal $\delta$, comparative
Maxents can obtain relatively optimal performance compared with the
performance using other $\delta$'s. The parametric configuration is
listed in Table~\ref{tabMaxentConfig}.

\begin{table}[h]
\centering
\begin{tabular}[ht]{l|p{5cm}}
\hline Category & Detailed Configurations \\
\hline Generating Times & $r = 10$ \\
\hline \#Bin & $m_{Syn, Dexter, ISOLET} = 100$ \\
 &$m_{SAT} = 50$, $m_{Sonar} = 30$\\
\hline Sample Size & $n = 10\cdot m$ \\
\hline Sampling Times & $s = 20$ \\
\hline \#Certain Constraints & $k = 0.2 \times m \ , \ k=0.05 \times m $ \\
\hline Threshold parameter& $\delta = 6e-4$ chosen from $[1e-4, 1.6e-3]$ \\
\hline
\end{tabular}
\caption{Parametric Configuration in Maxent Framework}
\label{tabMaxentConfig}
\end{table}

\subsubsection{Results}
\label{maxent_rst} We employ the parametric configuration in
Table~\ref{tabMaxentConfig} to run every Maxent. The mean
performance scores w.r.t. JS-Divergence and Expected Log Loss,
averaged on $r\times s$ sampling distributions, are summarized in
Tables~\ref{tabSynJSME}, \ref{tabSynLLME},~\ref{tabRWJSME}
and~\ref{tabRWLLME}. In addition to the performance scores, we also
give the best value and worst value w.r.t. the two evaluation
criteria. Finally, the overall performance of each algorithm,
averaged over all synthesized and all real-world datasets,  are
shown in Table~\ref{PS_prior_JS} and Table~\ref{PS_prior_LH},
respectively.

\begin{table*}[ht]
 {\hfill{}
\begin{tabular}[h]{c|c|c|c}
\hline \backslashbox{Method}{Data} & $U(0,1)$ & $|N(0,1^2)|$ & $N(3,1^2)$\\
\hline
\hline      Sample              & 0.5983/0.8088    & 0.7005/0.8838   & 0.0000/0.0000\\
\hline      F-$l^2_2$-TEBC      & 0.8639/0.9572    & 0.8695/0.9966   & 0.9793/{\bf 1.0000}\\
\hline      F-JSD-TEBC          & 0.9985/0.9999    & 0.9983/0.9992   & 0.9919/0.9863\\
\hline      F-ML-TEBC           & {\bf 1.0000}/{\bf 1.0000}    & {\bf 1.0000}/{\bf 1.0000}   & {\bf 1.0000}/0.9940\\
\hline      B-$l^2_2$-TEBC      & 0.8630/0.9571    & 0.8690/0.9962   & 0.9759/0.9955\\
\hline      B-JSD-TEBC          & 0.9978/0.9993    & 0.9981/0.9990   & 0.9875/0.9803\\
\hline      B-ML-TEBC           & 0.9992/0.9994    & 0.9997/0.9997   & 0.9955/0.9879\\
\hline      SME                 & 0.0000/0.0622    & 0.0231/0.0660   & 0.3774/0.6114\\
\hline      $l_1$-SUMMET        & 0.0017/0.0000    & 0.0000/0.0000   & 0.6376/0.6428\\
\hline      $l_1$-PLUMMET       & 0.0773/0.0566    & 0.0901/0.0577   & 0.4280/0.6299\\
\hline      $l^2_2$-SEB         & 0.8392/0.7936    & 0.7190/0.8508   & 0.3398/0.0889\\
\hline      JSD-SEB             & 0.9985/0.8494    & 0.9309/0.9296   & 0.3867/0.1172\\
\hline      ML-SEB              & 0.8410/0.8496    & 0.9332/0.9299   & 0.3921/0.1176\\
\hline
\hline      Best JS Value       & 0.0120/0.0143    & 0.0133/0.0156   & 0.0091/0.0113\\
\hline      Worst JS Value      & 0.0262/0.0338    & 0.0281/0.0340   & 0.0182/0.0190\\
\hline
\hline \backslashbox{Method}{Data} & ${\chi ^2}(10)$ & ${\beta}(3,6)$ & $B(30,0.2)$\\
\hline
\hline      Sample              & 0.0000/0.0000  & 0.0000/0.0000  & 0.0000/0.0000\\
\hline      F-$l^2_2$-TEBC      & {\bf 1.0000}/{\bf 1.0000}  & {\bf 1.0000}/{\bf 1.0000}  & {\bf 1.0000}/{\bf 1.0000}\\
\hline      F-JSD-TEBC          & 0.8736/0.7449  & 0.9243/0.8088  & 0.9443/0.8759\\
\hline      F-ML-TEBC           & 0.8830/0.7560  & 0.9347/0.8190  & 0.9534/0.8859\\
\hline      B-$l^2_2$-TEBC      & 0.9963/0.9955  & 0.9967/0.9958  & 0.9957/0.9957\\
\hline      B-JSD-TEBC          & 0.8704/0.7409  & 0.9212/0.8046  & 0.9396/0.8709\\
\hline      B-ML-TEBC           & 0.8798/0.7518  & 0.9314/0.8147  & 0.9486/0.8807\\
\hline      SME                 & 0.4469/0.7508  & 0.3993/0.5139  & 0.4686/0.7009\\
\hline      $l_1$-SUMMET        & 0.7528/0.7562  & 0.6077/0.4997  & 0.7484/0.7088\\
\hline      $l_1$-PLUMMET       & 0.5371/0.7649  & 0.4859/0.5135  & 0.5623/0.7070\\
\hline      $l^2_2$-SEB         & 0.3620/0.0802    & 0.3635/0.0795   & 0.3299/0.0951\\
\hline      JSD-SEB             & 0.4414/0.1373    & 0.4634/0.1414   & 0.3656/0.1217\\
\hline      ML-SEB              & 0.4450/0.1373    & 0.4674/0.1419   & 0.3689/0.1218\\
\hline
\hline      Best JS Value       & 0.0109/0.0128  & 0.0111/0.0129  & 0.0099/0.0113\\
\hline      Worst JS Value      & 0.0186/0.0190  & 0.0184/0.0189  & 0.0191/0.0184\\
\hline \hline \multicolumn{4}{c}{results w.r.t certain constraints'
number $k = 0.2 \times m (left) \ and \ 0.05 \times m (right) $}
\end{tabular}}\hfill{}
\caption{Performance Score (w.r.t. JS-divergence) of different
Maxents on Synthesized Datasets} \label{tabSynJSME}
\end{table*}

\begin{table*}[ht]
 {\hfill{}
\begin{tabular}[h]{c|c|c|c}
\hline \backslashbox{Method}{Data} & $U(0,1)$ & $|N(0,1^2)|$ & $N(3,1^2)$ \\
\hline
\hline      Sample              & 0.8539/0.7594    & 0.9179/0.8291    & 0.3802/0.0547   \\
\hline      F-$l^2_2$-TEBC      & 0.9331/0.9871    & 0.9190/{\bf 1.0000}    & {\bf 1.0000}/{\bf 1.0000}    \\
\hline      F-JSD-TEBC          & 0.9989/0.9987    & 0.9993/0.9840    & 0.9914/0.9371   \\
\hline      F-ML-TEBC           & {\bf 1.0000}/{\bf 1.0000}    & {\bf 1.0000}/0.9856    & 0.9982/0.9471   \\
\hline      B-$l^2_2$-TEBC      & 0.9293/0.9863    & 0.9191/0.9993    & 0.9977/0.9958  \\
\hline      B-JSD-TEBC          & 0.9986/0.9977    & 0.9992/0.9836    & 0.9884/0.9313  \\
\hline      B-ML-TEBC           & 0.9996/0.9989    & 0.9998/0.9851    & 0.9952/0.9413  \\
\hline      SME                 & 0.0000/0.1225    & 0.0000/0.0115    & 0.0000/0.0000  \\
\hline      $l_1$-SUMMET        & 0.7169/0.0719    & 0.7764/0.0000    & 0.8133/0.7731  \\
\hline      $l_1$-PLUMMET       & 0.5838/0.0000    & 0.6521/0.0192    & 0.5717/0.7539 \\
\hline      $l^2_2$-SEB         & 0.7553/0.5448    & 0.7943/0.4707   & 0.5039/0.0400 \\
\hline      JSD-SEB             & 0.9319/0.7853    & 0.9741/0.8590   & 0.6117/0.1514 \\
\hline      ML-SEB              & 0.9327/0.7855    & 0.9749/0.8594   & 0.6152/0.1518  \\
\hline
\hline      Best ELL Value      & 6.4128/6.4179    & 6.2935/6.3022    & 6.5928/6.6009 \\
\hline      Worst ELL Value     & 6.5864/6.4879    & 6.5226/6.3587    & 6.6583/6.6394 \\
\hline
\hline \backslashbox{Method}{Data} & ${\chi ^2}(10)$ & ${\beta}(3,6)$ & $B(30,0.2)$\\
\hline
\hline      Sample              & 0.4724/0.0936   & 0.6084/0.0911  & 0.7054/0.0236 \\
\hline      F-$l^2_2$-TEBC      & {\bf 1.0000}/{\bf 1.0000}   & {\bf 1.0000}/{\bf 1.0000}  & {\bf 1.0000}/{\bf 1.0000}   \\
\hline      F-JSD-TEBC          & 0.9239/0.7299   & 0.9642/0.7823  & 0.9770/0.8411 \\
\hline      F-ML-TEBC           & 0.9298/0.7408   & 0.9693/0.7930  & 0.9803/0.8529 \\
\hline      B-$l^2_2$-TEBC      & 0.9985/0.9967   & 0.9982/0.9962  & 0.9988/0.9959      \\
\hline      B-JSD-TEBC          & 0.9221/0.7262   & 0.9629/0.7784  & 0.9755/0.8360\\
\hline      B-ML-TEBC           & 0.9279/0.7369   & 0.9679/0.7889  & 0.9788/0.8477 \\
\hline      SME                 & 0.0000/0.0000   & 0.0000/0.6955  & 0.0000/0.7849 \\
\hline      $l_1$-SUMMET        & 0.9061/0.8648   & 0.8935/0.6993  & 0.9371/0.7977\\
\hline      $l_1$-PLUMMET       & 0.6539/0.8531   & 0.7555/0.7008  & 0.8527/0.7923\\
\hline      $l^2_2$-SEB         & 0.5335/0.0147    & 0.6291/0.0000   & 0.7546/0.0000 \\
\hline      JSD-SEB             & 0.6897/0.1989    & 0.7776/0.1984   & 0.8075/0.1278 \\
\hline      ML-SEB              & 0.6917/0.1990    & 0.7793/0.1989   & 0.8085/0.1279  \\
\hline
\hline      Best ELL Value      & 6.5460/6.5533   & 6.5405/6.5491  & 6.5916/6.5834 \\
\hline      Worst ELL Value     & 6.6129/6.5873   & 6.6267/6.5820  & 6.7347/6.6178 \\
\hline \multicolumn{4}{c}{results w.r.t certain constraints' number
$k = 0.2 \times m (left) \ and \ 0.05 \times m (right) $}
\end{tabular}}\hfill{}
\caption{Performance Scores (w.r.t. Expected Log Loss) of different
Maxents on Synthesized Datasets} \label{tabSynLLME}
\end{table*}

\begin{table*}[ht]
 {\hfill{}
\begin{tabular}[h]{c|c|c|c|c}
\hline \backslashbox{Method}{Data} & Dexter & Statlog & ISOLET & Sonar \\
\hline
\hline      Sample              & 0.3852/0.4350    & 0.8212/0.8968    & 0.5107/0.5693    & 0.3323/0.3176\\
\hline      F-$l^2_2$-TEBC      & 0.9415/{\bf 1.0000}    & 0.8000/0.9010    & 0.8261/0.9997    & 0.9513/{\bf 1.0000} \\
\hline      F-JSD-TEBC          & 0.9955/0.7596    & 0.9996/0.9995    & 0.9966/0.9565    & 0.9956/0.8274\\
\hline      F-ML-TEBC           & {\bf 1.0000}/0.7631    & 0.9999/{\bf 1.0000}    & {\bf 1.0000}/0.9598    & {\bf 1.0000}/0.8328 \\
\hline      B-$l^2_2$-TEBC      & 0.9395/0.9983    & 0.7998/0.9008    & 0.8250/{\bf 1.0000}    & 0.9476/0.9948      \\
\hline      B-JSD-TEBC          & 0.9952/0.7591    & 0.9996/0.9995    & 0.9951/0.9550    & 0.9942/0.8239 \\
\hline      B-ML-TEBC           & 0.9996/0.7624    & {\bf 1.0000}/0.9999    & 0.9984/0.9582    & 0.9984/0.8289 \\
\hline      SME                 & 0.0000/0.0697    & 0.0000/0.0065    & 0.0000/0.0691    & 0.0000/0.0737 \\
\hline      $l_1$-SUMMET        & 0.3694/0.0000    & 0.2099/0.0000    & 0.1384/0.0000    & 0.0962/0.0000 \\
\hline      $l_1$-PLUMMET       & 0.1956/0.0819    & 0.2588/0.1281    & 0.0867/0.0842    & 0.1148/0.1567 \\
\hline      $l^2_2$-SEB         & 0.5313/0.4094    & 0.7533/0.7945    & 0.5821/0.5497    & 0.5605/0.3684      \\
\hline      JSD-SEB             & 0.8874/0.5702    & 0.9853/0.9716    & 0.8675/0.6704    & 0.8117/0.5095 \\
\hline      ML-SEB              & 0.8910/0.5709    & 0.9863/0.9719    & 0.8704/0.6709    & 0.8144/0.5092 \\
\hline
\hline      Best JS Value       & 0.0144/0.0152    & 0.0132/0.0148    & 0.0132/0.0145    & 0.0134/0.0142\\
\hline      Worst JS Value      & 0.0213/0.0213    & 0.0337/0.0304    & 0.0227/0.0228    & 0.0201/0.0197\\
\hline \multicolumn{5}{c}{results w.r.t certain constraints' number
$k = 0.2 \times m (left) \ and \ 0.05 \times m (right) $}
\end{tabular}}\hfill{}
\caption{Performance Score (w.r.t. JS-divergence) of different
Maxents on Real-world Datasets} \label{tabRWJSME}
\end{table*}

\begin{table*}[ht]
 {\hfill{}
\begin{tabular}[h]{c|c|c|c|c}
\hline \backslashbox{Method}{Data} & Dexter & Statlog & ISOLET & Sonar \\
\hline
\hline      Sample              & 0.9023/0.4710    & 0.9811/0.9884    & 0.8680/0.7702    & 0.9696/0.9648 \\
\hline      F-$l^2_2$-TEBC      & 0.9463/{\bf 1.0000}    & 0.9351/0.9572    & 0.8621/0.9833    & 0.9838/{\bf 1.0000} \\
\hline      F-JSD-TEBC          & 0.9991/0.7678    & 0.9998/0.9999    & 0.9986/0.9974    & 0.9997/0.9954 \\
\hline      F-ML-TEBC           & {\bf 1.0000}/0.7714    & 0.9999/{\bf 1.0000}    & {\bf 1.0000}/{\bf 1.0000}    & {\bf 1.0000}/0.9959 \\
\hline      B-$l^2_2$-TEBC      & 0.9451/0.9970    & 0.9342/0.9569    & 0.8618/0.9919    & 0.9823/0.9992      \\
\hline      B-JSD-TEBC          & 0.9990/0.7671    & 0.9998/0.9998    & 0.9981/0.9963    & 0.9995/0.9952 \\
\hline      B-ML-TEBC           & 0.9998/0.7706    & {\bf 1.0000}/0.9999    & 0.9995/0.9988    & 0.9998/0.9955 \\
\hline      SME                 & 0.0000/0.5313    & 0.5737/0.4980    & 0.0000/0.5079    & 0.7454/0.5920 \\
\hline      $l_1$-SUMMET        & 0.9470/0.4908    & 0.8644/0.6724    & 0.8252/0.4840    & 0.8715/0.8095 \\
\hline      $l_1$-PLUMMET       & 0.8330/0.5286    & 0.0000/0.0000   & 0.6469/0.0000    & 0.0000/0.0000 \\
\hline      $l^2_2$-SEB         & 0.7848/0.0000    & 0.9028/0.8834    & 0.7169/0.5678    & 0.9333/0.9203      \\
\hline      JSD-SEB             & 0.9755/0.5619    & 0.9962/0.9920    & 0.9557/0.8100    & 0.9889/0.9740 \\
\hline      ML-SEB              & 0.9761/0.5626    & 0.9964/0.9920    & 0.9567/0.8103    & 0.9891/0.9740 \\
\hline
\hline      Best ELL Value      & 6.3275/6.3600    & 5.0336/5.0862    & 6.0809/6.2262    & 4.6626/4.6838\\
\hline      Worst ELL Value     & 6.5260/6.3918    & 5.7561/5.4920    & 6.2318/6.2896    & 5.3435/5.1230 \\
\hline \multicolumn{5}{c}{results w.r.t certain constraints' number
$k = 0.2 \times m (left) \ and \ 0.05 \times m (right) $}
\end{tabular}}\hfill{}
\caption{Performance Scores (w.r.t. Expected Log Loss) of different
Maxents on Real-world Datasets}\label{tabRWLLME}
\end{table*}

\begin{table}[ht]
\begin{center}
\begin{tabular}[h]{c|c|c|c}
\hline Algorithms &         Synthesized &     Real-world &      Average\\
\hline      Sample              & 0.2164/0.2821 & 0.5123/0.5547      & 0.3644/0.4184 \\
\hline      F-$l^2_2$-TEBC      & 0.9521/{\bf 0.9923} & 0.8797/{\bf 0.9752}      & 0.9159/{\bf 0.9837}  \\
\hline      F-JSD-TEBC          & 0.9551/0.9025 & 0.9968/0.8858      & 0.9760/0.8941  \\
\hline      F-ML-TEBC           & {\bf 0.9618}/0.9091 & {\bf 0.9999}/0.8889      & {\bf 0.9809}/0.8990  \\
\hline      B-$l^2_2$-TEBC      & 0.9495/0.9893 & 0.8780/0.9735      & 0.9137/0.9814      \\
\hline      B-JSD-TEBC          & 0.9524/0.8992 & 0.9960/0.8844      & 0.9742/0.8918  \\
\hline      B-ML-TEBC           & 0.9590/0.9057 & 0.9991/0.8874      & 0.9791/0.8965  \\
\hline      SME                 & 0.2859/0.4508 & 0.0000/0.0547      & 0.1429/0.2528 \\
\hline      $l_1$-SUMMET        & 0.4580/0.4346 & 0.2035/0.0000      & 0.3307/0.2173  \\
\hline      $l_1$-PLUMMET       & 0.3634/0.4549 & 0.1640/0.1127      & 0.2637/0.2838  \\
\hline      $l^2_2$-SEB         & 0.4658/0.3313 & 0.6068/0.5305      & 0.5363/0.4309       \\
\hline      JSD-SEB             & 0.5712/0.3828 & 0.8879/0.6804      & 0.7296/0.5316  \\
\hline      ML-SEB              & 0.5746/0.3830 & 0.8905/0.6807      & 0.7326/0.5319  \\
\hline \multicolumn{4}{c}{results w.r.t certain constraints' number
$k = 0.2 \times m \ / \ k=0.05 \times m  $}
\end{tabular}
\end{center}
\caption{Overall Performance Score evaluated by JS-Divergence for
Experiment Results in Section~\ref{priorExp}}\label{PS_prior_JS}
\end{table}
\begin{table}[ht]
\begin{center}
\begin{tabular}[h]{c|c|c|c}
\hline Algorithms &         Synthesized &     Real-world &      Average\\
\hline      Sample              & 0.6564/0.3086   & 0.9302/0.7986     & 0.7933/0.5536  \\
\hline      F-$l^2_2$-TEBC      & 0.9753/{\bf 0.9978}   & 0.9318/0.9851     & 0.9536/{\bf 0.9915} \\
\hline      F-JSD-TEBC          & 0.9758/0.8789   & 0.9993/0.9401     & 0.9875/0.9095\\
\hline      F-ML-TEBC           & {\bf 0.9796}/0.8866   & {\bf 0.9999}/0.9418     & {\bf 0.9898}/0.9142  \\
\hline      B-$l^2_2$-TEBC      & 0.9736/0.9950   & 0.9309/{\bf 0.9862}     & 0.9522/0.9906 \\
\hline      B-JSD-TEBC          & 0.9744/0.8756   & 0.9991/0.9396     & 0.9868/0.9076  \\
\hline      B-ML-TEBC           & 0.9782/0.8831   & 0.9998/0.9412     & 0.9890/0.9122  \\
\hline      SME                 & 0.0000/0.2690   & 0.3298/0.5323     & 0.1649/0.4007  \\
\hline      $l_1$-SUMMET        & 0.8406/0.5345   & 0.8770/0.6142     & 0.8588/0.5743 \\
\hline      $l_1$-PLUMMET       & 0.6783/0.5199   & 0.3699/0.1321     & 0.5241/0.3260  \\
\hline      $l^2_2$-SEB         & 0.6618/0.1783   & 0.8345/0.5929     & 0.7481/0.3856 \\
\hline      JSD-SEB             & 0.7988/0.3868   & 0.9791/0.8345     & 0.8889/0.6106  \\
\hline      ML-SEB              & 0.8004/0.3871   & 0.9796/0.8347     & 0.8900/0.6109 \\
\hline \multicolumn{4}{c}{results w.r.t certain constraints' number
$k = 0.2 \times m  \ / \ k=0.05 \times m  $}
\end{tabular}
\end{center}
\caption{Overall Performance Score evaluated by Expected Log Loss
for Experiment Results in Section~\ref{priorExp}}\label{PS_prior_LH}
\end{table}

When $k=0.2 \times m$, in experimental results on synthesized
datasets, all TEBC Maxents outperform comparative Maxents in most
cases. From Table \ref{PS_prior_JS}, we can observe that, on
average, F-ML-TEBC and B-ML-TEBC are the best two models among TEBC
Maxents. The superiority of TEBC Maxents is clearly shown in Tables
\ref{tabSynJSME} and \ref{PS_prior_JS}, under the JS-divergence
measure. This superiority also holds under expected log loss
measure, which is demonstrated in Tables~\ref{tabSynLLME}
and~\ref{PS_prior_LH}. As for real-world datasets, we can still draw
the same conclusion that TEBC Maxents show their advantages over the
others from Tables \ref{tabRWJSME} and \ref{tabRWLLME}. Like the
results on synthesized datasets, F-ML-TEBC and B-ML-TEBC still show
their robustness and become the best two models in the average sense
from Tables \ref{PS_prior_JS} and \ref{PS_prior_LH}.


When $k=0.05 \times m$, the amount of prior information is actually
reduced. In this case, it is also clearly demonstrated that all TEBC
Maxents outperform comparative Maxents on average. Particularly, we
can observe that F-$l^2_2$-TEBC and B-$l^2_2$-TEBC become the most
stable and effective, somewhat differing from their performance in
the case where $k=0.2 \times m$.

We would like to mention that in our experiments, standard Maxent
and $l_1$-PLUMMET did not perform well. In many cases, using the
sampling distribution directly (denoted as Sample) can outperform
these two Maxents, especially in the experiment with real-world
datasets. One of the causes is that when we fix a unified $\delta$
to all the uncertain constraints, it is often in a dilemma: If a
small $\delta$ is used, the feasible region might be far from the
underlying real distribution, and hence the optimization procedure
can only converge to a poor solution; If a large $\delta$ is used,
there might be a great chance to underfit the sample. Compared with
standard Maxent and $l_1$-PLUMMET, $l_1$-SUMMET (using the
selective-update strategy) is relatively stable. However,
$l_1$-SUMMET with any unified $\delta$ can not perform as well as
TEBC Maxents. Note that there is no practical guidance to determine
different $\delta$ for different uncertain constraints. Even though
this guidance exists, it is often a prohibitively-complicated task
to find the optimal $\delta$ for each uncertain constraint. This
indeed reflects the implementation complexity of the existing
Maxents and highlights the advantage of the parameter-free
characteristic of TEBC Maxents. The idea is also supported by the
experiment result of SEB Maxents, which achieve better performance
than SME and GME on average although are still less effective than
TEBCs.

%

In summary, TEBC Maxents show their effectiveness and stability, as
demonstrated in result tables, especially in Tables
\ref{PS_prior_JS} and \ref{PS_prior_LH}. We would like to stress
that TEBC Maxents are also easy to implement since only certain
constraints and a single TEB constraint are involved. This can help
to demonstrate our previous theoretical justification on TEB. In
addition, note that the performance of SEB-based models can
consistently outperform SME, $l_1$-SUMMET and $l_1$-PLUMMET, though
the SEB correction is indeed not exact. This observation indicates
that the framework of our generalized Maxent proposed in
Section~\ref{model} has gains in itself.

\subsection{TEB-Lidstone vs. Comparative Lidstone and Good-Turing Estimators}
\label{noPriorExp} When the reliable prior information is not
available, Lidstone and Good-Turing estimators are often used in
many applications since they are often effective enough, and more
efficient than Maxent. This set of experiments is constructed to
verify the advantage of TEB-Lidstone (F-Lidstone and B-Lidstone)
over the involved Lidstone and Good-Turing estimators.

\subsubsection{Lidstone and Good-Turing Estimators}
\label{Lidstones} In this subsection, we evaluate the effectiveness
of F-Lidstone and B-Lidstone estimators which have been described in
Section~\ref{connection}. Various other Lidstone models, i.e.,
Laplace estimator (Laplace) and Expected Likelihood Estimator (ELE),
serve as comparative algorithms. Motivated by TEB-Lidstone, we also
derive a Lidstone estimator from SEB in a similar way to
TEB-Lidstone. To identify the rate of probability correction $f$,
the SEB constraint (Formula~\ref{SEB}) is used as a guidance instead
of the TEB constraint, so that the Shannon entropy of the resulting
distribution $S\left[ {\bar P^{(m)} } \right]$ is closest to the sum
of the sampling Shannon entropy $S\left[ {\widehat P_n^{(m)} }
\right]$ and the Shannon entropy bias $\Delta S$. The resulting
Lidstone estimator is referred as SEB-Lidstone.

In addition, some Good-Turing estimators, i.e., the simplest
Good-Turing estimator (SimplestGT), Simple Good-Turing
(SGT)~\citep{Gale95} and a low complexity diminishing attenuation
estimator (LC-DAE) with some asymptotic guarantee of
performance~\citep{Orlitsky03}, are also involved in comparative
experiments. Because Good-Turing estimators do not assume the bin
number $m$ is given, they only assign a probability sum to all the
zero bins w.r.t the sample. In our case in which $m$ is provided,
the sum is uniformly distributed to all these zero bins.

\subsubsection{Results}
\label{Lidstone_rst} We employ the same parameter setting in
Table~\ref{tabMaxentConfig} but ignore the parameters in constraints
of Maxent, and then run F-Lidstone and B-Lidstone estimators as well
as the involved comparative models on the synthesized and real-world
datasets. The mean performance scores w.r.t. JS-Divergence and
Expected Log Loss over generating times $r$ and sampling times $s$
for each dataset are summarized in Table~\ref{tabSynJS} to
Table~\ref{tabRWLL}. The overall performance of each algorithm on
synthesized and real-world datasets is also demonstrated in
Table~\ref{PS_no_prior_JS} and Table~\ref{PS_no_prior_LH}.

\begin{table*}[ht]
 {\hfill{}
\begin{tabular}[h]{c|c|c|c|c|c|c}
\hline \backslashbox{Method}{Data} & $U(0,1)$ & $|N(0,1^2)|$ & $N(3,1^2)$ & ${\chi ^2}(10)$ & ${\beta}(3,6)$ & $B(30,0.2)$\\
\hline
\hline      Sample &        0.0000      &  0.0000       &  0.0000       &  0.0000       &  0.0000       &  0.0000 \\
\hline      SimplestGT &    0.0916      &  0.1851       &  0.0062       &  0.0446       &  0.0679       &  0.0219 \\
\hline      Laplace &       0.9429      &  {\bf 1.0000}       &  0.4823       &  0.6518       &  0.6521       &  0.5082 \\
\hline      ELE &           0.6318      &  0.7745       &  0.2741       &  0.3744       &  0.3785       &  0.2876 \\
\hline      SGT &           0.4896      &  0.3537       &  0.3754       &  0.3235       &  0.4152       &  0.3642 \\
\hline      LC-DAE &        0.5680      &  0.7053       &  0.2532       &  0.5771       &  0.5202       &  0.4032 \\
\hline      B-Lidstone &    0.9999      &  0.9566      &  0.9958       &  0.9954       &  0.9961       &  0.9953 \\
\hline      F-Lidstone &    {\bf 1.0000}      &  0.9573       &  {\bf 1.0000}      &  {\bf 1.0000}        &  {\bf 1.0000}      &  {\bf 1.0000} \\
\hline      SEB-Lidstone &  0.7768      &  0.8065      &  0.5755       &  0.6500       &  0.6363       &  0.5882 \\
\hline
\hline      Best JS Value   & 0.0151    & 0.0154    & 0.0114    & 0.0132    & 0.0131    & 0.0118 \\
\hline      Worst JS Value  & 0.0181    & 0.0177    & 0.0183    & 0.0187    & 0.0186    & 0.0187 \\
\hline
\end{tabular}}\hfill{}
\caption{Performance Score (w.r.t. JS-divergence) of different
Lidstone and Good-Turing Estimators on Synthesized
Datasets}\label{tabSynJS}
\end{table*}

\begin{table*}[ht]
{\hfill{}
\begin{tabular}[h]{c|c|c|c|c|c|c}
\hline \backslashbox{Method}{Data} & $U(0,1)$ & $|N(0,1^2)|$ & $N(3,1^2)$ & ${\chi ^2}(10)$ & ${\beta}(3,6)$ & $B(30,0.2)$\\
\hline
\hline      Sample &        0.0000      &  0.0000       &  0.0000       &  0.0000       &  0.0000       &  0.0000 \\
\hline      SimplestGT &    0.1475      &  0.2419       &  0.0228       &  0.0748       &  0.1074       &  0.0423 \\
\hline      Laplace &       0.9042      &  {\bf 1.0000}       &  0.4965       &  0.6732       &  0.6715       &  0.5313 \\
\hline      ELE &           0.5798      &  0.6886       &  0.2860       &  0.3967       &  0.3995       &  0.3076 \\
\hline      SGT &           0.4857      &  0.3881       &  0.3861       &  0.3487       &  0.4391       &  0.3817 \\
\hline      LC-DAE &        0.7220      &  0.8708       &  0.3367       &  0.6309       &  0.5951       &  0.4599 \\
\hline      B-Lidstone &    0.9985      &  0.9258       &  0.9957       &  0.9957       &  0.9963       &  0.9956\\
\hline      F-Lidstone &    {\bf 1.0000}      &  0.9273      &  {\bf 1.0000}       &  {\bf 1.0000}        &  {\bf 1.0000}       &  {\bf 1.0000} \\
\hline      SEB-Lidstone &  0.7260      &  0.7349      &  0.5865        &  0.6689       &  0.6542       &  0.6071 \\
\hline
\hline      Best ELL Value  & 6.4366   & 6.2912    & 6.6035    & 6.5505    & 6.5494    & 6.5945 \\
\hline      Worst ELL Value & 6.4539   & 6.3051    & 6.6364    & 6.5781    & 6.5769    & 6.6275 \\
\hline
\end{tabular}}\hfill{}
\caption{Performance Scores (w.r.t. Expected Log Loss) of different
Lidstone and Good-Turing Estimators on Synthesized
Datasets}\label{tabSynLL}
\end{table*}

\begin{table}[ht]
 {\hfill{}
\begin{tabular}[h]{c|c|c|c|c}
\hline \backslashbox{Method}{Data} & Dexter & Statlog & ISOLET & Sonar \\
\hline
\hline      Sample &        0.0000          &  0.4568       &   0.0000      &  0.0000  \\
\hline      SimplestGT &    0.2774          &  0.3649       &   0.1241      &  0.1712  \\
\hline      Laplace &       {\bf 1.0000}          &  0.4363       &   0.8874      &  0.8967 \\
\hline      ELE &           0.6277          &  0.9992       &   0.5832      &  0.5480  \\
\hline      SGT &           0.3740          &  0.4704       &   0.3137      &  0.5455  \\
\hline      LC-DAE &        0.8305          &  0.0000       &   0.6002      &  0.6556  \\
\hline      B-Lidstone &    0.9099          &  {\bf 1.0000}       &   0.9972      &  0.9912 \\
\hline      F-Lidstone &    0.9124          &  0.9986       &   {\bf 1.0000}     &  {\bf 1.0000}  \\
\hline      SEB-Lidstone &  0.7658          &  0.9204       &   0.7668     &  0.7172  \\
\hline
\hline      Best JS Value   & 0.0148   & 0.0152    & 0.0150   & 0.0144 \\
\hline      Worst JS Value  & 0.0184   & 0.0171    & 0.0183   & 0.0182 \\
\hline
\end{tabular}}\hfill{}
\caption{Performance Score (w.r.t. JS-divergence) of different
Lidstone and Good-Turing Estimators on Real-world
Datasets}\label{tabRWJS}
\end{table}

\begin{table}[ht]
 {\hfill{}
\begin{tabular}[h]{c|c|c|c|c}
\hline \backslashbox{Method}{Data} & Dexter & Statlog & ISOLET & Sonar \\
\hline
\hline      Sample&         0.0000      &  0.1702       &   0.0000      &  0.0000  \\
\hline      SimplestGT &    0.3568      &  0.0000       &   0.1795      &  0.2092  \\
\hline      Laplace &       {\bf 1.0000}      &  0.8697       &   0.9138      &  0.9062  \\
\hline      ELE &           0.6363      &  {\bf 1.0000}       &   0.5827      &  0.5631  \\
\hline      SGT &           0.4422      &  0.2000       &   0.3584      &  0.5509  \\
\hline      LC-DAE &        0.9297      &  0.3987       &   0.7471      &  0.7839  \\
\hline      B-Lidstone &    0.9153      &  0.9195       &   0.9970     &  0.9911  \\
\hline      F-Lidstone &    0.9177      &  0.9157       &   {\bf 1.0000}     &  {\bf 1.0000}  \\
\hline      SEB-Lidstone &  0.7727      &  0.6445       &   0.7562     &  0.7240  \\
\hline
\hline      Best ELL Value  & 6.3450   & 4.9949   & 6.4027   & 4.7125 \\
\hline      Worst ELL Value & 6.3659   & 5.0012   & 6.4207   & 4.7338 \\
\hline
\end{tabular}}\hfill{}
\caption{Performance Scores (w.r.t. Expected Log Loss) of different
Lidstone and Good-Turing Estimators on Real-world
Datasets}\label{tabRWLL}
\end{table}

\begin{table}[ht]
\begin{center}
\begin{tabular}[h]{c|c|c|c}
\hline Algorithms &         Synthesized &     Real-world &      Average\\
\hline SimplestGT &          0.0695   &      0.2344 &       0.1520\\
\hline Laplace &             0.7062   &      0.8051 &       0.7556\\
\hline ELE &                 0.4535   &      0.6896 &       0.5715\\
\hline SGT &                 0.3869   &      0.4259 &       0.4064\\
\hline LC-DAE &              0.5045   &      0.5216 &       0.5130\\
\hline B-Lidstone &          0.9899   &      0.9746 &       0.9822\\
\hline F-Lidstone &          {\bf 0.9928}   &      {\bf 0.9777} &       {\bf 0.9853}\\
\hline SEB-Lidstone &        0.6722   &      0.7925 &       0.7324\\
\hline
\end{tabular}
\end{center}
\caption{Overall Performance Score evaluated by JS-Divergence for
Experiment Results in
Section~\ref{noPriorExp}}\label{PS_no_prior_JS}
\end{table}

\begin{table}[ht]
\begin{center}
\begin{tabular}[h]{c|c|c|c}
\hline Algorithms &         Synthesized &     Real-world &      Average\\
\hline SimplestGT &          0.1061   &      0.1864 &       0.1462\\
\hline Laplace &             0.7128   &      0.9224 &       0.8176\\
\hline ELE &                 0.4431   &      0.6955 &       0.5693\\
\hline SGT &                 0.4049   &      0.3879 &       0.3964\\
\hline LC-DAE &              0.6025   &      0.7149 &       0.6587\\
\hline B-Lidstone &          0.9846   &      0.9557 &       0.9702\\
\hline F-Lidstone &          {\bf 0.9878}   &      {\bf 0.9583} &       {\bf 0.9731}\\
\hline SEB-Lidstone &        0.6629   &      0.7243 &       0.6936\\
\hline
\end{tabular}
\end{center}
\caption{Overall Performance Score evaluated by Expected Log Loss
for Experiment Results in
Section~\ref{noPriorExp}}\label{PS_no_prior_LH}
\end{table}

In experimental results on synthesized datasets, it can be observed
that F-Lidstone and B-Lidstone estimators, especially F-Lidstone,
outperform the other models in most cases. Even in the worst case,
they are still more effective than most of the other models. Hence,
F-Lidstone and B-Lidstone are on average the best performing among
all estimators.

For real-world datasets, we can still come to the same conclusion
that F-Lidstone and B-Lidstone outperform the others on average.
Although Laplace and ELE could achieve the optimal performance on
Dexter and Statlog datasets, the effectiveness of F-Lidstone and
B-Lidstone is just slightly lower but their performance on the other
datasets, especially ISOLET, is much better than that of the other
estimators. Further, it can be observed that in some cases the
performance scores evaluated by JS-Divergence and Expected Log Loss
are not consistent with each other. The phenomenon is probably due
to the incompleteness of these evaluation criteria, which could only
partially reflect the similarity between the resulting distribution
and the underlying real distribution. In this sense, the more stable
the algorithm is in different similarity criteria, the more
effective it could be considered to be. F-Lidstone and B-Lidstone
are also optimal in this way.

In summary, when Bayesian-TEB and Frequentist-TEB are applied to the
Lidstone framework, the resulting B-Lidstone and F-Lidstone
estimators could achieve excellent performance in the expected
sense, compared with common Lidstone and Good-Turing estimators.
Hence, it can be concluded that TEBs do make sense and can benefit
the Lidstone framework.

\section{Conclusions and Future Work}
\label{ConFut}

This paper proposes the closed-form formulae on the expected Tsallis
entropy bias (TEB) under Frequentist and Bayesian frameworks. TEBs
give the quantities on the difference between the expected Tsallis
entropy of sampling distributions and the Tsallis entropy of the
underlying real distribution. It is exact in the sense of
unbiasedness and consistency, and hence naturally entails a
quantitative re-interpretation of the Maxent principle. In other
words, TEBs quantitatively give the answer to the question: Why we
should choose the distribution with maximum entropy. We further use
TEBs in Maxent and Lidstone frameworks, and both of them show
promising results on synthesized and real-world datasets.

In using maximum entropy approach for density estimation, a key
challenge lies in the dilemma of uncertain constraints selection:
Inappropriate choices may easily cause serious overfitting or
underfitting problems. To deal with the challenge, a family of TEBC
Maxents, namely $l_2^2$-TEBC, JSD-TEBC and ML-TEBC, are proposed in
this paper. Instead of using uncertain constraints selected
empirically, the proposed models let its Tsallis entropy converge to
the underlying real distribution by compensating expected Tsallis
entropy bias, while ensure the resulting distribution to resemble
the sampling distribution w.r.t. $l_2^2$ norm, JS-divergence or
Maximum Likelihood. Hence, the resulting distributions are optimal
in the expected sense, w.r.t. the above three similarity criteria.

The family of TEBC Maxents is a natural generalization of standard
Maxent. The important difference between TEBC Maxents and standard
Maxent is that, the constraints of the former can be derived from
reliable prior information (certain constraints) or analytical
analysis (TEB constraint), while the latter also has to involve
uncertain constraints demanding empirically parametric selection. It
turns out that TEBC Maxents become parameter-free in this sense.
Furthermore, the analytically established TEB constraint can be
effective to depress overfitting or underfitting, since it can force
the resulting distribution to approach the real one by matching
their entropies.

In addition to the Maxent framework, we also demonstrate that there
is a natural connection between TEB and another widely used
estimator, Lidstone estimator. Specifically, TEB can analytically
identify the adaptive rate of probability correction in Lidstone
framework. As a result, TEB-Lidstone estimators (F-Lidstone and
B-Lidstone) have been developed.

In the future, several extra theoretical issues are worth
considering. Firstly, the TEB results might be developed in terms of
other $q$ indexes of Tsallis entropy. The unbiased and consistent
results w.r.t $q=1$ is of special interests since Tsallis entopy is
equivalent to Shannon entropy in this case. It can be expected that
these extended results offer more complete criteria to further solve
the overfitting and underfitting. For instance, as an extreme case,
if we can give $m$ independent Frequentist-TEB results, the
possibility of overfitting and underfitting can be, in principle,
ruled out and hence a task of $m$-nomial distribution estimation
could become determined. However, other numerical procedures should
be devised in order to globally solve the resulting model
integrating TEB constraints w.r.t. $q\neq 2$, since the new TEB
constraints could be non-convex. Secondly, it is also interesting to
develop Bayesian-TEB results w.r.t other Bayesian priors. Finally,
we have observed that the two criteria of the estimation quality,
i.e., JS-divergence and the expected log loss, occasionally give
inconsistent evaluation results. Hence, it is helpful to develop
more sophisticated metrics to evaluate the performance of density
estimation.

\begin{acknowledgements}
The authors would like to thank editors and anonymous reviewers in
advance for their constructive comments. This work is supported in
part by the Natural Science Foundation of China (Grant 60603027),
the HK Research Grants Council (project no. CERG PolyU 5211/05E and
5217/07E), the UK's Engineering and Physical Sciences Research
Council (EPSRC, grants EP/E002145/1 and EP/F014708/1), the Natural
Science Foundation project of Tianjin, China (grant 09JCYBJC00200),
and MSRA Internet Services Theme Project, 2008.
\end{acknowledgements}

\newpage

\appendix
\section*{Appendix A. Standard Deviation of $E_{n,p}^{(m)} \left( T
\right)$}
\begin{Theorem}
\label{thrm:four} Given the uniform probability metric over the
$\mathbb P^{(m)}$, the standard deviation of $E_{n,p}^{(m)} \left( T
\right)$ denoted by $STD_n^{(m)} \left( T \right)$, then
\begin{equation}
\label{thrm:four-one} STD_n^{(m)} \left( T \right) = \frac{2}{{\sqrt
{(m - 1) \cdot (m + 2) \cdot (m + 3)} }}E_n^{(m)} \left( T \right)
\end{equation}
\begin{proof}
Combining the definition of standard deviation with the integral
operator $L^{(m - 1)}$ defined by Formula~\ref{thrm:two-three}, we
have:
\[
STD_n^{(m)} \left( T \right) = \sqrt {\frac{1}{{Z^{(m - 1)} }}L^{(m
- 1)} \left\{ {\left[ {E_{P,n}^{(m)} \left( T \right) - E_n^{(m)}
\left( T \right)} \right]^2 } \right\}}
\]
which could be further transformed into£º
\begin{equation}
\label{thrm:four-two}
\begin{gathered}
STD_n^{(m)} \left( T \right) = \sqrt{\frac{1}{{Z^{(m - 1)} }} \cdot
L^{(m - 1)} \left\{ {\left[ {E_{P,n}^{(m)} \left( T \right)}
\right]^2 } \right\} - \left[ {E_n^{(m)} \left( T \right)}
\right]^2}
\end{gathered}
\end{equation}
Recall that $E_{P,n}^{(m)} \left( T \right)$ has been re-expressed
in Formula~\ref{thrm:two-two}, and hence it can be verified that
\[
\begin{gathered}
\left[ {E_{P,n}^{(m)} \left( T \right)} \right]^2  = \frac{{4\left(
{n - 1} \right)^2 }}{{n^2 }}\left( {\sum\nolimits_{i = 1}^{m - 1}
{p_i }  - \sum\nolimits_{i = 1}^{m - 1} {p_i^2 }} { -
\sum\nolimits_{i = 1}^{m - 1} {\sum\nolimits_{j = i + 1}^{m - 1}
{p_i  \cdot p_j } } } \right)^2 \hfill
\end{gathered}
\]
Let $U^{(m - 1)}$ denote the set of all product terms in $\left(
{\sum\nolimits_{i = 1}^{m - 1} {p_i }  - \sum\nolimits_{i = 1}^{m -
1} {p_i^2 }  - \sum\nolimits_{i = 1}^{m - 1} {\sum\nolimits_{j = i +
1}^{m - 1} {p_i  \cdot p_j } } } \right)^2$. Then the calculation of
$\left[ {E_{P,n}^{(m)} \left( T \right)} \right]^2$ is reduced to
solve the closed-form $U^{(m - 1)}$ w.r.t. integral operator $L^{(m
- 1)}$.

Due to the symmetry of integral domain, the properties of $L^{(m -
1)}$ in Proposition~\ref{thrm:two} could be generalized as the
followings:
\begin{enumerate}
\item[(a)] $L^{(m - 1)} (p_i^r ) = L^{(m - 1)} (p_j^r ),1 \le i,j \le m - 1,r
\in \{ 2,3,4\}$
\item[(b)] $L^{(m - 1)} (p_i  \cdot p_j^r ) = L^{(m - 1)} (p_k  \cdot p_l^r ),1
\le i,j,k,l \le m - 1,i \ne j,k \ne l,r \in \{ 1,2,3\}$
\item[(c)] $L^{(m - 1)} (p_i^2  \cdot p_j^2 ) = L^{(m - 1)} (p_k^2  \cdot
p_l^2),1 \le i,j \le m - 1,i \ne j,k \ne l$
\item[(d)] $L^{(m - 1)} (p_i  \cdot p_j  \cdot p_k^r ) = L^{(m - 1)} (p_u  \cdot
p_v  \cdot p_w^r ), 1 \le i,j,k,u,v,w \le m - 1,i \ne j \ne k,u \ne
v \ne w,r \in \{ 2,3\}$
\item[(e)] $L^{(m - 1)} (p_i  \cdot p_j  \cdot p_k  \cdot p_l ) = L^{(m - 1)}
(p_u  \cdot p_v  \cdot p_w  \cdot p_x ), 1 \le i,j,k,l,u,v,w,x \le m
- 1,i \ne j \ne k \ne l,u \ne v \ne w \ne x$
\end{enumerate}

Based on the above properties, we can obtain a partition of $U^{(m -
1)}$: First, we partition $U^{(m - 1)}$ into five different sets
subject to the formal constraints of (a)-(e); Second, we further
partition each set into subsets subject to different $r$ values (if
$r$ is involved in the formal definition of a set). After the above
two steps of decomposition, the final partition includes 10 parts
and could be represented as below:
\[
\begin{gathered}
{\rm{Partition}}\left[ {U^{(m)} } \right]{\rm{ = }} \left\{ {\left\{
{p_i^2 } \right\},\left\{ {p_i^3 } \right\},\left\{ {p_i^4 }
\right\},\left\{ {p_i p_j } \right\},\left\{ {p_i p_j^2 } \right\}}
\right., \hfill \\ \left.{\left\{ {p_i p_j^3 } \right\},\left\{
{p_i^2 p_j^2 } \right\},\left\{ {p_i p_j p_k } \right\},\left\{ {p_i
p_j p_k^2 } \right\},\left\{ {p_i p_j p_k p_l } \right\}} \right\}
\hfill
\end{gathered}
\]

It can checked that the terms in each part give the same result with
respect the integral operator $L^{(m - 1)}$.

Therefore, if the general term formula of each part and their term
numbers could be worked out, the general term formula of $\left[
{E_{P,n}^{(m)} \left( T \right)} \right]^2$ follows directly.

The following equations could be checked:
\[
\begin{gathered}
L^{(m - 1)} \left( {p_i^2 } \right) = \frac{{2!}}{{\left( {m + 1}
\right)!}} \hfill \\
L^{(m - 1)} \left( {p_i^3 } \right) = \frac{{3!}}{{\left( {m + 2}
\right)!}} \hfill \\
L^{(m - 1)} \left( {p_i^4 } \right) = \frac{{3!}}{{\left( {m + 3}
\right)!}} \hfill \\
L^{(m - 1)} \left( {p_i  \cdot p_j } \right) = \frac{1}{{\left( {m +
1} \right)!}} \hfill \\
L^{(m - 1)} \left( {p_i  \cdot p_j^2 } \right) = \frac{{2!}}{{\left(
{m + 2} \right)!}} \hfill \\
L^{(m - 1)} \left( {p_i  \cdot p_j^3 } \right) = \frac{{3!}}{{\left(
{m + 3} \right)!}} \hfill \\
L^{(m - 1)} \left( {p_i^2  \cdot p_j^2 } \right) = \frac{4}{{\left(
{m + 3} \right)!}} \hfill \\
L^{(m - 1)} \left( {p_i  \cdot p_j  \cdot p_k } \right) =
\frac{1}{{\left( {m + 2} \right)!}} \hfill \\
L^{(m - 1)} \left( {p_i  \cdot p_j  \cdot p_k^2 } \right) =
\frac{{2!}}{{\left( {m + 3} \right)!}} \hfill \\
L^{(m - 1)} \left( {p_i  \cdot p_j  \cdot p_k  \cdot p_l } \right) =
\frac{1}{{\left( {m + 3} \right)!}} \hfill \\
\end{gathered}
\]

In addition, the general term formula $N^{(m - 1)} ( \cdot )$ of the
term number in each part is given by
\[
\begin{gathered}
N^{(m - 1)} (p_i^2 ) = m - 1 \hfill \\
N^{(m - 1)} (p_i^3 ) =  - 2(m - 1) \hfill \\
N^{(m - 1)} (p_i^4 ) = m - 1 \hfill \\
N^{(m - 1)} (p_i p_j ) = (m - 1)(m - 2) \hfill \\
N^{(m - 1)} (p_i p_j^2 ) =  - 4(m - 1)(m - 2) \hfill \\
N^{(m - 1)} (p_i p_j^3 ) = 2(m - 1)(m - 2) \hfill \\
N^{(m - 1)} (p_i^2 p_j^2 ) = \frac{{3 \cdot (m - 1)(m - 2)}}{2} \hfill \\
N^{(m - 1)} (p_i p_j p_k ) =  - (m - 1)(m - 2)(m - 3) \hfill \\
N^{(m - 1)} (p_i p_j p_k^2 ) = 2(m - 1)(m - 2)(m - 3) \hfill \\
N^{(m - 1)} (p_i p_j p_k p_t ) = \frac{{(m - 1)(m - 2)(m - 3)(m -
4)}}{4} \hfill \\
\end{gathered}
\]

Replace $\left[ {E_{P,n}^{(m)} \left( T \right)} \right]^2$ by the
two sets of general term formulae, and we can get
\begin{equation}
\label{thrm:four-three}
\begin{gathered}
\frac{1}{{Z^{(m - 1)} }} \cdot L^{(m - 1)} \left\{ {\left[
{E_{P,n}^{(m)} \left( T \right)} \right]^2 } \right\} \hfill \\ =
\frac{1}{{Z^{(m)} }} \cdot \frac{{4\left( {n - 1} \right)^2 }}{{n^2
}}\sum\limits_{\alpha  \in {\rm{Partition}}\left[ {U^{(m)} }
\right]} {L^{(m - 1)} \left( \alpha  \right) \cdot N^{(m - 1)}
(\alpha )} \hfill \\
= \frac{{\left( {n - 1} \right)^2 }}{{n^2 }}\frac{{\left( {m^3  +
2m^2  - 5m + 2} \right)}}{{(m + 1)(m + 2)(m + 3)}} \hfill
\end{gathered}
\end{equation}

Substitute Formula~\ref{thrm:four-three} into
Formula~\ref{thrm:four-two}, and then the following equation holds:
\[
STD_n^{(m)} \left( T \right) = \frac{{n - 1}}{{n \cdot }}\sqrt
{\frac{{\left( {m^3  + 2m^2  - 5m + 2} \right)}}{{(m + 1)(m + 2)(m +
3)}} - \frac{{(m - 1)^2 }}{{(m + 1)^2 }}}
\]
After the simplification, we obtain:
\[
\begin{gathered}
STD_n^{(m)} \left( T \right) = \frac{2}{{\sqrt {(m - 1) \cdot (m +
2) \cdot (m + 3)} }}\cdot \frac{{(n - 1)}}{n}\frac{{\left( {m - 1}
\right)}}{{(m + 1)}}
\end{gathered}
\]
Recall that $E_n^{(m)} \left( T \right) = \frac{{(n - 1) \cdot (m -
1)}}{{n \cdot (m + 1)}}$, and hence Formula~\ref{thrm:four-one} is
proved.
\end{proof}
\end{Theorem}
It is clear that, if $n \approx m$, $STD_n^{(m)} \left( T \right)
\approx \frac{2}{{\sqrt m }}\frac{{m - 1}}{{n \cdot (m + 1)}}$.
Recall that $\frac{{m - 1}}{{n \cdot \left( {m + 1} \right)}}$ is
uniform Bayesian-TEB. Hence, the estimation of uniform Bayesian-TEB
is relatively stable in the case of inadequate sampling.

\vskip 0.2in

\bibliographystyle{plainnat}
\bibliography{myBibFile}   

\end{document}